\documentclass[journal]{IEEEtran}

\usepackage[pdftex]{graphicx}
\graphicspath{{pdf/}{jpeg/}}
\DeclareGraphicsExtensions{.pdf,.jpeg,.png}

\usepackage{multicol}
\usepackage{tabularx, booktabs, ragged2e}
\usepackage{times}
\usepackage{comment}
\usepackage{subcaption}

\usepackage{bbm}
\usepackage{soul}
\usepackage{url}
\usepackage[utf8]{inputenc}
\usepackage{amsmath}
\usepackage{multirow}
\usepackage{adjustbox}
\usepackage{balance}
\usepackage[table,xcdraw,dvipsnames]{xcolor}
\usepackage{balance}
\usepackage[]{algorithm2e}
\usepackage{tabularx, booktabs, ragged2e}
\usepackage{rotating}

\usepackage{enumitem}     
\usepackage[acronym]{glossaries}
\usepackage{acronym}
\usepackage{ltablex, enumitem, makecell, float, textcomp}
\usepackage{array,colortbl}
\usepackage{bibunits}
\usepackage{comment}
\usepackage{pifont}
\usepackage{lipsum}

\usepackage{multirow}
\usepackage[pagebackref=true,breaklinks=true,letterpaper=true,colorlinks,bookmarks=false, citecolor=blue]{hyperref}

\usepackage{xspace}

\newcommand{\etal}{\textit{et al}. }
\newcommand{\ie}{\textit{i}.\textit{e}., }
\newcommand{\eg}{\textit{e}.\textit{g}., }
\acrodef{gnn}[GNN]{graphical neural network}
\acrodef{fg}[FG]{IEEE International Conference on Automatic Face and Gesture Recognition}
\newcommand{\cf}{\textit{c}.\textit{f}., }

\acrodef{ml}[ML]{machine learning}
\acrodef{sota}[SOTA]{state-of-the-art}

\acrodef{hog}[HOG]{histogram of gradients}
\acrodef{ss}[SS]{sister-sister}
\acrodef{bb}[BB]{brother-brother}
\acrodef{sibs}[SIBS]{brother-sister}

\acrodef{tfidf}[TF-IDF]{term frequency-inverse document frequency learning}

\acrodef{fs}[FS]{father-son}
\acrodef{ms}[MS]{mother-son}
\acrodef{fd}[FD]{father-daughter}

\acrodef{md}[MD]{mother-daughter}
\acrodef{gfgs}[GFGS]{grandfather-grandson}
\acrodef{gmgs}[GMGS]{grandmother-grandson}
\acrodef{gfgd}[GFGD]{grandfather-granddaughter}

\acrodef{gmgd}[GMGD]{grandmother-granddaughter}

\acrodef{sdm}[SDM]{signal detection model}
\acrodef{roc}[ROC]{receiver operating characteristic}
\acrodef{nmse}[NMSE]{Normalized Mean Square Error}
\acrodef{det}[DET]{Detection Error Trade-off}
\acrodef{tp}[TP]{true-positive}
\acrodef{tn}[TN]{true-negative}
\acrodef{ap}[AP]{average precision}
\acrodef{ae}[AE]{autoencoder}

\acrodef{bce}[BCE]{Binary Cross Entropy}
\acrodef{tpir}[TPIR]{true-positive identification rate}
\acrodef{frir}[FRIR]{false-reject identification rate}
\acrodef{fpir}[FRIR]{false-positive identification rate}

\acrodef{fn}[FN]{false-negative}
\acrodef{frr}[FRR]{false-reject rate}
\acrodef{fnr}[FNR]{false-negative rate}
\acrodef{fp}[FP]{false-positive}

\acrodef{fpr}[FPR]{\ac{fp} rate}
\acrodef{tpr}[TPR]{true-positive rate}

\acrodef{fiw}[FIW]{\textit{Families In the Wild}}
\acrodef{tsk}[TSKIN]{\textit{Tri-Subject Kinship}}

\acrodef{kfw}[KinFaceW]{\textit{Kin-Faces in the Wild}}
\acrodef{kfvw}[KFVW]{\textit{KinFaceW Videos}}

\acrodef{rfiw}[RFIW]{\textit{Recognizing Families In the Wild}}

\acrodef{nn}[NN]{neural network}

\acrodef{cnn}[CNN]{Convolutional Neural Network}
\acrodef{caae}[CAEE]{conditional adversarial \ac{ae}}
\acrodef{lut}[LUT]{Look-Up-Table}
\acrodef{fr}[FR]{face recognition}
\acrodef{gan}[GAN]{generative adversarial network}
\acrodef{dae}[DAE]{denoising \ac{ae}}

\acrodef{svm}[SVM]{Support Vector Machine}
\acrodef{mid}[MID]{Member ID}
\acrodef{fid}[FID]{Family ID}
\acrodef{pid}[PID]{Photo ID}
\acrodef{roc}[ROC]{receiver operating characteristic}
\acrodef{nrml}[NRML]{Neighborhood Repulsed Metric Learning}

\acrodef{fsp}[FSP]{`From same photograph'}
\definecolor{ao(english)}{rgb}{0.0, 0.5, 0.0}

\acrodef{dkmr}[DKMR]{Deep Kinship Matching and Recognition}

\acrodef{lflkin}[LFL-KIN]{Linear Feature Learning}
\acrodef{hsl}[HSL]{Heterogeneous Similarity Learning}

\acrodef{sml}[SML]{Support Vector Data Description-based metric learning}
\acrodef{msml}[MSML]{multi-view SML}
\acrodef{lm3l}[LM$^3$L]{large-margin multi-metric learning}

\acrodef{svdd}[SVDD]{Support Vector Data Description}

\acrodef{tfidf}[TF-IDF]{term frequency-inverse document frequency}
\acrodef{nlp}[NLP]{Natural Language Processing}
\acrodef{map}[MAP]{mean average precision}

\newcommand{\xmark}{\ding{56}}%
\newcommand{\checkc}{\ding{51}}%

\newacronym{ml}{ML}{machine learning}
\newacronym{fr}{FR}{facial recognition}
\newacronym{fv}{FV}{facial verification}

\newacronym{cnn}{CNN}{convolutional neural network}
\newacronym{knn}{K-NN}{\emph{K} nearest neighbors}

\newacronym{nn}{NN}{neural network}
\newacronym{mtcnn}{MTCNN}{\emph{multi-task \gls{cnn}}}

\newacronym{gan}{GAN}{generative adversarial network}
\newacronym{se}{SE}{\emph{Squeeze-and-Excitation}}
\newacronym{d}{$D$}{discriminator}
\newacronym{g}{$G$}{generator}
\newacronym{dbvae}{DB-VAE}{Debiasing Variational Autoencoder}

\newacronym{lut}{LUT}{Look-Up-Table}
\newacronym{soa}{SOTA}{state-of-the-art}

\newacronym{fiw}{FIW}{Families In the Wild}
\newacronym{lfw}{LFW}{Labeled Faces in the Wild}
\newacronym{bfw}{BFW}{Balanced Faces in the Wild}
\newacronym{rfw}{RFW}{Racial Faces in the Wild:}
\newacronym{dp}{DemogPairs}{Demographic Pairs}
\newacronym{itwcc}{ITWCC}{Wild Child Celebrity}
\newacronym{dif}{D\emph{i}F}{Diversity in Faces}

\newacronym{m}{M}{\textit{Male}}
\newacronym{f}{F}{\textit{Female}}
\newacronym{a}{A}{\textit{Asian}}
\newacronym{b}{B}{\textit{Black}}
\newacronym{i}{I}{\textit{Indian}}
\newacronym{w}{W}{\textit{White}}
\newacronym{af}{AF}{\textit{Asian}-\textit{Female}}
\newacronym{am}{AM}{\textit{Asian}-\textit{Male}}
\newacronym{bf}{BF}{\textit{Black}-\textit{Female}}
\newacronym{bm}{BM}{\textit{Black}-\textit{Male}}
\newacronym{if}{IF}{\textit{Indian}-\textit{Female}}
\newacronym{im}{IM}{\textit{Indian}-\textit{Male}}
\newacronym{wf}{WF}{\textit{White}-\textit{Female}}
\newacronym{wm}{WM}{\textit{White}-\textit{Male}}

\newacronym{mlp}{MLP}{multi-layered perceptron}

\newacronym{fd}{FD}{Face Discrimination}
\newacronym{bb}{BB}{bounding box}

\newacronym{sdm}{SDM}{signal detection model}
\newacronym{roc}{ROC}{receiver operating characteristic}
\newacronym{nmse}{NMSE}{Normalized Mean Square Error}
\newacronym{det}{DET}{Detection error trade-off}
\newacronym{tp}{TP}{true-positive}
\newacronym{fp}{FP}{false-positive}

\newacronym{tpir}{TPIR}{true-positive identification rate}
\newacronym{frir}{FRIR}{false-reject identification rate}
\newacronym{fpir}{FRIR}{false-positive identification rate}

\newacronym{da}{DA}{domain adaptation}
\newacronym{fn}{FN}{false-negative}
\newacronym{frr}{FRR}{false-reject rate}
\newacronym{fnr}{FNR}{false-negative rate}

\newacronym{fpr}{FPR}{false-positive rate}

\newacronym{tpr}{TPR}{true-positive rate}
\newacronym{tn}{TN}{true-negative}
\newacronym{tnr}{TNR}{true-negative rate}
\newacronym{eer}{EER}{Equal Error Rate}

\newacronym{cs}{CS}{Cosine Similarity}

\newacronym{ita}{ITA}{individual typology angle}
\newacronym{fst}{FST}{Fitzpatrick skin type}
\newacronym{mi}{MI}{melanin index}

\newacronym{lime}{LIME}{Local Interpretable Model-Agnostic Explanations}
\newacronym{nas}{NAS}{Neural Architecture Search}
\newacronym{gapf}{GAPF}{Generative Adversarial Privacy and Fairness}

\definecolor{Gray}{gray}{0.85}
\definecolor{LightCyan}{rgb}{0.88,1,1}

\newcolumntype{a}{>{\columncolor{Gray}}c}
\newcolumntype{b}{>{\columncolor{white}}c}
\renewcommand{\arraystretch}{1.4}

\newcommand{\vo}{\vec{o}\@ifnextchar{^}{\,}{}}

\newcommand{\vx}{\vec{x}\@ifnextchar{^}{\,}{}}
\usepackage{amssymb}%

\def\colorModel{hsb}
\usepackage{babel}

\usepackage{tikz}
\usepackage{cite}

\usepackage[T1]{fontenc}

\usepackage[table]{xcolor}
\usepackage{collcell}
\usepackage{hhline}
\usepackage{pgf}
\usepackage{multirow}

\usepackage{pgfplots}

\usepackage[TS1,T1]{fontenc}
\usepackage{fourier, heuristica}

\usepackage{caption}
\DeclareCaptionFont{black}{\color{black}}
\newcommand{\foo}{\color{black}\makebox[0pt]{\textbullet}\hskip-0.5pt\vrule width 1pt\hspace{\labelsep}}

\usepackage{MnSymbol} 

\pgfplotsset{compat=1.8}
\usepgfplotslibrary{statistics}
\def\colorModel{hsb} 

\newcommand\ColCell[1]{
  \pgfmathparse{#1<50?1:0}  
    \ifnum\pgfmathresult=0\relax\color{white}\fi
  \pgfmathsetmacro\compA{0}      
  \pgfmathsetmacro\compB{#1/100} 
  \pgfmathsetmacro\compC{1}      
  \edef\x{\noexpand\centering\noexpand\cellcolor[\colorModel]{\compA,\compB,\compC}}\x #1
  } 
\newcolumntype{E}{>{\collectcell\ColCell}m{0.55cm}<{\endcollectcell}}  

\definecolor{tip}{rgb}{0.0, 0.0, 0.0}
\newcommand{\tip}[1]{\color{tip}{#1}~\color{black}}

\definecolor{tipr}{rgb}{0.0, 0.0, 0.0}
\newcommand{\tipr}[1]{\color{tipr}{#1}~\color{black}}

\definecolor{dark}{rgb}{0.23921569, 0.16470588, 0.1372549}
\definecolor{tana}{rgb}{0.38823529, 0.24313725, 0.16862745}
\definecolor{tanb}{rgb}{0.49019608, 0.33333333, 0.22352941}
\definecolor{interma}{rgb}{0.59215686, 0.43921569, 0.28627451}
\definecolor{intermb}{rgb}{0.70980392, 0.56470588, 0.44705882}
\definecolor{lighta}{rgb}{0.78431373, 0.66666667, 0.54901961}
\definecolor{lightb}{rgb}{0.91372549, 0.74901961, 0.6627451 }
\definecolor{verylight}{rgb}{0.95686275,0.84313725, 0.70588235}

\usepackage{array}
\newcolumntype{L}[1]{>{\raggedright\let\newline\\\arraybackslash\hspace{0pt}}m{#1}}
\newcolumntype{C}[1]{>{\centering\let\newline\\\arraybackslash\hspace{0pt}}m{#1}}
\newcolumntype{R}[1]{>{\raggedleft\let\newline\\\arraybackslash\hspace{0pt}}m{#1}}

\usepackage{url}

 \usepackage[pdftex]{thumbpdf}

\begin{document}
\title{Balancing Biases and Preserving Privacy on Balanced Faces in the Wild}

\author{Joseph P Robinson,~\IEEEmembership{Member,~IEEE,}
        Can Qin,~\IEEEmembership{Student Member,~IEEE,} 
        Yann Henon, \\Samson Timoner,~\IEEEmembership{Member,~IEEE,} 
        and~Yun Fu,~\IEEEmembership{Fellow,~IEEE}

\thanks{Manuscript accepted 16 May 2023.}

}
\markboth{Journal of IEEE TIP}%
{How to Use the IEEEtran \LaTeX \ Templates}

\maketitle

\IEEEtitleabstractindextext{%
\glsresetall
\begin{abstract}
There are demographic biases present in current \gls{fr} models. To measure these biases across different ethnic and gender subgroups, we introduce our \gls{bfw} dataset. This dataset allows for the characterization of FR performance per subgroup. We found that relying on a single score threshold to differentiate between \emph{genuine} and \emph{imposters} sample pairs leads to suboptimal results. Additionally, performance within subgroups often varies significantly from the \emph{global} average. Therefore, specific error rates only hold for populations that match the validation data. To mitigate imbalanced performances, we propose a novel domain adaptation learning scheme that uses facial features extracted from state-of-the-art neural networks. This scheme boosts the average performance and preserves identity information while removing demographic knowledge. Removing demographic knowledge prevents potential biases from affecting decision-making and protects privacy by eliminating demographic information. We explore the proposed method and demonstrate that subgroup classifiers can no longer learn from features projected using our domain adaptation scheme. For access to the source code and data, please visit \href{https://github.com/visionjo/facerec-bias-bfw}{https://github.com/visionjo/facerec-bias-bfw}.
\end{abstract}
\glsresetall

}

\IEEEdisplaynontitleabstractindextext

\IEEEpeerreviewmaketitle

\ifCLASSOPTIONcompsoc
\IEEEraisesectionheading{\section{Introduction}\label{sec:introduction}}
\else
\section{Introduction}
\label{sec:introduction}
\fi

\glsresetall

\IEEEPARstart{A}{s} machine learning \gls{ml} becomes more integrated into our daily lives, interest in concepts like bias, fairness, and the ethical implications of using this technology grows~\cite{10.1007/978-3-030-13469-3_68, anne2018women, wang2018racial}. As we rely more on \gls{ml} to assist with everyday tasks, it becomes increasingly critical to address biased and unfair algorithms ~\cite{lazo2020towards, raji2019actionable}. Systems deployed for sensitive tasks require thorough examination, with biometrics~\cite{drozdowski2020demographic}: \gls{fr} being a prime example. We propose a test bed to evaluate \gls{fr} fairly.

\begin{figure}
    \centering
    \includegraphics[width=.65\linewidth]{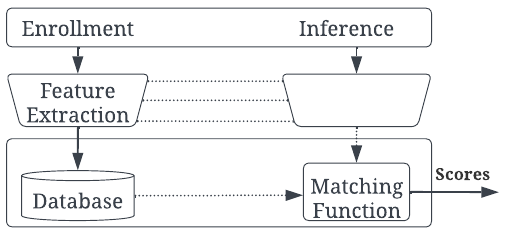}
    \caption{{\textbf{Generic \gls{fr} system.} \emph{Enrollment:} {encode face images and store them} in a database. \emph{Inference:} extract features from a test face and match them to those in the database to produce scores.}}
    \label{fig:generic-fr}
\end{figure}

 \tipr{Researchers and practitioners often use \glspl{cnn} or transformer models} to map face features to a vector. These \gls{fr} models are typically trained on data to learn to encode faces in a space where those of the same identity are minimally separated while those of different identities are furthest apart. The \gls{fr} model then extracts features (Fig.~\ref{fig:generic-fr}) to store in a database with labels. 
 
\begin{figure}
\glsunset{fnr}
\glsunset{fpr}
\glsunset{tpr}
    \centering
    \includegraphics[width=.85\linewidth]{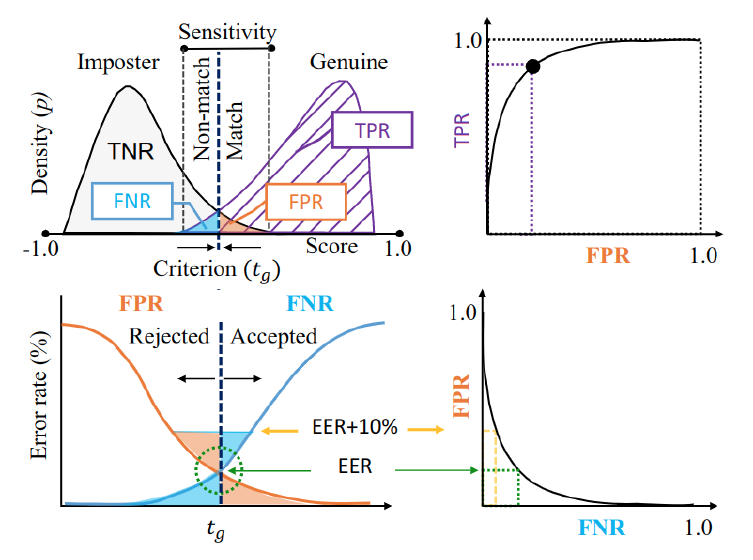}
    \caption{\textbf{Depiction of biometrics.} The score distribution shows how a single threshold ($t_g$) affects FR. Top-left illustrates the threshold. Top-right shows how \gls{tpr} changes with \gls{fpr}. Bottom-left and bottom-right show the trade-off between \gls{fnr} and \gls{fpr}. The range is metric-dependent (\eg cosine similarity $-1\leq t_g \leq 1$).} 
    \label{fig:metrics}
    \glsreset{fnr}\glsreset{fpr}\glsreset{tpr}
\end{figure}

 \tip{Subsequently, during inference, one compares the features of a test face to faces stored during enrollment to determine a match.} \tip{An optimal} threshold (\ie $\theta$) serves as the decision boundary to compare the similarity score $s$ of a pair of unseen \tipr{faces} to predict the pair-wise class label (\ie \emph{genuine} or \emph{imposter}). Ideally, the face features of {true} pairs yield scores that satisfy criterion $s \ge \theta$~\cite{deng2019arcface, liu2017sphereface, wang2018additive, wang2018cosface}: $\theta$ serves as a trade-off parameter \tip{to control} the \gls{fpr} and \gls{fnr}  (Fig.~\ref{fig:metrics}).

The adverse effects of a \tipr{single} threshold are threefold: 
\begin{enumerate}
    \item Evaluation sets typically have imbalanced distributions similar to the training, so the majority dominates the performance rating.
    \item Score ranges for genuine vary across demographics; true face pairs from the underrepresented subgroups tend to score lower.
    \item The optimal scores per subgroup vary, meaning a single, global threshold is only optimal if set and tested on a single sub-population (Fig.~\ref{fig:detcurves}).
\end{enumerate}

Throughout this work, we use the term \emph{global} to refer to values averaged across all demographics, in contrast to \emph{subgroup-specific}, which refers to a particular demographic. To address the issue of imbalanced data (\ie item \textbf{(1)} mentioned above), we propose the \gls{bfw} dataset to measure subgroup biases in \gls{fr}. BFW provides a fair evaluation of FR systems by considering demographic-specific performance. We can now understand the performance gap in facial features with \gls{soa} \glspl{cnn}. We then suggest a mechanism to eliminate prejudice and level out performance ratings across demographics while improving accuracy. Specifically, we preserve identity information and remove demographic knowledge from the features. This feature adaptation scheme addresses items \textbf{2}-\textbf{3}.



\tipr{A byproduct of the proposed is to preserve privacy.} The learned features in the {lower-dimensional} space contain less knowledge of the {subgroups,} \tip{disallowing the extraction of} ethnicity and gender information from \tip{the enrolled facial features.} Protecting user data is {valuable,} and reduces the chance of malicious or even unintended bias~\cite{bowyer2004face}.

The contributions {are listed as follows.}

\begin{itemize}
    \item We demonstrate a bias in \glspl{cnn} with our \gls{bfw} dataset. We added another attribute representing facial skin tones. \tipr{The raw data, face embeddings, and meta-data are on IEEEDataPort.}\footnote{\href{https://ieee-dataport.org/documents/balanced-faces-wild}{https://ieee-dataport.org/documents/balanced-faces-wild}.}
    \item We propose a feature learning scheme that {de-biases} face {features} {and} balances performances across subgroups, \tipr{increasing performance.}
    \item We minimize subgroup information in features -- a byproduct of the proposed work is the reduction of subgroup-based knowledge {to address privacy concerns and avoid other potential biases.}
    \item We draw attention to the challenging samples the suggested de-biasing scheme overcomes. 
\end{itemize}

The paper is organized as follows. We review related work (Section~\ref{sec:relatedworks}). Then, we go over constructing the \gls{bfw} database (Section~\ref{subsec:data}). We introduce the proposed method (Section~\ref{sec:proposed}). Then, the settings and results of the experiments are covered (Section~\ref{sec:experimental}). Finally, we discuss the next steps (Section~\ref{sec:conclusions}).

\section{Related Work}\label{sec:relatedworks}
\glsreset{fr}
\subsection{Bias in facial recognition}

Automatic \gls{fr} based on deep learning dates back to 2014, when Taigman~\etal~\cite{taigman2014deepface} first proposed using a \gls{cnn} for recognition, which has seen significant improvements nearly annually.
The \gls{soa} continues to improve, with layer and network types evolving (\eg transformers~\cite{ren2022shunted}). For more in-depth surveys, see~\cite{guo2019survey, masi2018deep}, and ~\cite{khalil2020investigating}.

\tip{Recent research focuses on reducing bias in automatic face understanding~\cite{buolamwini2018gender, 250171}.
 Some focus on the changes in performance (\ie biases) of} \emph{soft attributes} \tipr{like} gender~\cite{muthukumar2018understanding}, ethnicity, age~\cite{drozdowski2020demographic}, or other \tipr{traits~\cite{Terhost2021bias}.} \tipr{Others explore methods to measure bias using generative modeling.} For example, Balakrishnan~\etal~\cite{balakrishnan2020towards} train a generator to manipulate \tipr{the latent space to alter attributes such as skin tone, hair length,} and hair color. Georgopoulos~\etal~\cite{georgopoulos2020enhancing} generate faces \tipr{of} various ages {to} augment training data. {Muthukumar~\etal~\cite{9025567} study the effects of color tone on gender classification by recoloring faces in images across a spectrum, \tipr{\ie from lighter to darker.} Others \tipr{use knowledge distillation (\cf~\cite{dhar2021distill, liu2021rectifying}). Gong~\etal~\cite{gong2021mitigating} base subgroups on facial skin tones.} This paper focuses on the common one-to-one \gls{fv} setting.


%
\begin{table}[!b]
\caption{\textbf{Major milestones for deep learning in \gls{fr}.}}
\renewcommand\arraystretch{1.4}\arrayrulecolor{black}
\begin{tabular}{@{\,}r <{\hskip 2pt} !{\foo} >{\raggedright\arraybackslash}p{7cm}}
\toprule
\addlinespace[1.5ex]
2014 & Taigman~\etal~\cite{taigman2014deepface} proposed using a \gls{cnn} (\ie a Siamese network~\cite{chopra2005learning}) to do \gls{fv}.\\
2015 & Parkhi~\etal~\cite{Parkhi15} increased model size and trained with triplet-loss on more faces.\\
2016 & Wen~\etal~\cite{wen2016a} combined a softmax with their proposed center-loss that overcomes the burdens of sampling negative pairs.\\
2017 & Liu~\etal~\cite{liu2017sphereface} recalled that facial images exist on a manifold~\cite{4587670} and modified softmax (\ie SphereFace) to learn features on a hypersphere manifold optimally separated by the geodesic distance.\\
2018 & Deng~\etal~\cite{deng2019arcface} ArcFace trained with A-Softmax loss, a modified triplet loss to compare faces to $K$ positive and many more negative samples via angular margins, captured more global information than the previous sample-to-sample objective.\\
2019 & Duan~\etal~\cite{duan2019uniformface} A-Softmax loss and uniform loss improved the local knowledge and distribution on the hypersphere manifold.\\
2020 & Wang~\etal~\cite{wang2020masked} focused on better discriminating between faces wearing masks, for it had suddenly grown to have a great value.\\
2021 & Meng~\etal~\cite{meng2021magface} proposed MagFace, which assigned pairwise weights during training based on the difficulty (or ease). \newline Zhu~\etal~\cite{Zhu_2021_CVPR} leveraged larger quantities of higher-quality samples.\\
\end{tabular}
\label{fig:timelines} 
\end{table}

%

\tipr{Some researchers aim to characterize the amount of bias at the system level, including} gender~\cite{serna2020insidebias, albiero2020analysis, das2018mitigating}, ethnicity~\cite{wang2018racial}, age~\cite{srinivas2019face}, or multiple~\cite{nagpal2019deep,  acien2018measuring, gong2019debface, savchenko2019efficient, Nagpal_2020_CVPR_Workshops}. A recent \emph{European Conference on Computer Vision} (ECCV) challenge \tipr{encourages researchers to tackle bias in} ethnicity, gender, age, pose, and even with/without sunglasses~\cite{sixta2020fairface}. \tipr{Other methods add modalities (\eg profile information) to mitigate} bias~\cite{pena2020bias, pena2020faircvtest}. \tipr{Still, other works focus on the measurement of biases in \gls{fr} at different levels, including the system~\cite{acien2018measuring, serna2020insidebias}, templates~\cite{8987331}, scores~\cite{terhorst2020post}, and in pre-trained models~\cite{Sadeghi_2020_CVPR_Workshops}.} Wang~\etal~\cite{wang2020mitigating} introduce a reinforcement learning-based \tipr{race-balanced} network to find optimal margins for non-Caucasians. \tipr{Law~\etal~\cite{law2020designing} leverage HCI technology to detect bias semi-supervised by having a human in the loop. These works target the image space, whereas we target the features without the original model or face image.}

\tipr{Terhorst~\etal~\cite{terhorst2020post} and Cavazos~\etal~\cite{cavazos2020accuracy} recognize the same challenges shown in this work: the sensitivities in subgroups when a matching function is applied to generate a score from a pair of features vary across demographics. These  works normalize the scores to handle demographic-specific sensitivities to the average--an issue we highlight in~\cite{robinson2020face}.}

\begin{figure}[!t] 
	\centering    
 \glsunset{fpr}
  \glsunset{fnr}
    \glsunset{det}
	\includegraphics[width=.7\linewidth]{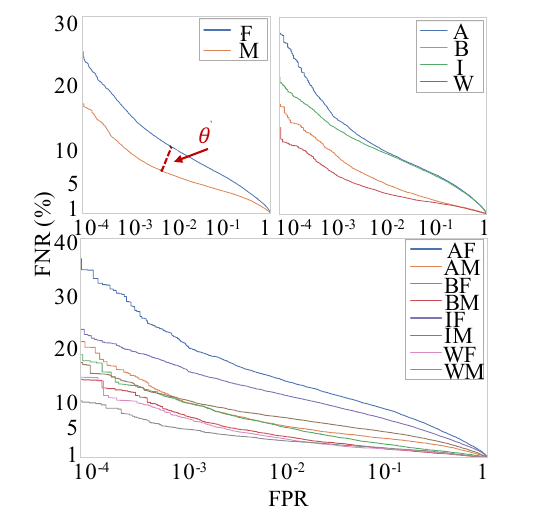}
 		\caption{{\textbf{\gls{det} curves.} \emph{Top-left}: per gender. \emph{Top-right}: per ethnicity. \emph{Bottom}: per subgroup (\ie combined). The dashed line shows about 2$\times$ difference in \gls{fpr} for the same threshold $\theta.$ \gls{fnr} is the match error count.}}
		\glsreset{det}\glsreset{fpr}
\label{fig:detcurves} 
\end{figure}

\subsection{Imbalanced data and data problems in FR}
\tipr{To effectively produce fair data distributions, one can under or over-sample sub-groups~\cite{drummond2003c4}. Alternatively, one can adjust learning costs per sub-group~\cite{Gong_2021_CVPR,yang2021ramface}. } Rudd~\etal~\cite{rudd2016moon} propose the mixed objective optimization network (MOON) architecture that learns to classify attributes of faces by treating each subgroup as a multi-task attribute (\ie a task per attribute). Cluster-based Large Margin Local Embedding (CLMLE)~\cite{huang2019deep} \tipr{samples} in the feature space \tipr{regularize} the models at the decision boundaries of underrepresented classes. Wang~\etal~\cite{wang2019balanced} change images by masking out aspects of humans that cause ``leakage'' of gender information to avoid biasing a set of labels describing a scene to a specific gender. \tipr{The less recent solution can be found in reviews~\cite{he2009learning,he2013imbalanced, krawczyk2016learning}.}

Drozdowski~\etal~\cite{drozdowski2020demographic} \tipr{claim} that the cohorts of concern in biometrics are demographic (\eg sex, age, and race), person-specific (\eg pose or expression~\cite{xu2020investigating}), and environmental (\eg camera-model, sensor size, illumination, and occlusion). Albiero~\etal~\cite{albiero2020does} and Gwilliam~\etal~\cite{Gwilliam_2021_ICCV} found that balanced data sets do not yield balanced results. {We study demographics' effect on} \gls{fv} by assessing demographic-specific {performances.} Our \gls{bfw} data resource allows us to analyze existing \gls{soa} deep \glspl{cnn} on different subgroups. {We provide practical insights showing that} experiments often report {misleading performance ratings that depend} on demographics.

Many researchers release large \gls{fr} datasets to match the capacity of modern-day deep models~\cite{guo2016ms, schroff2015facenet, Cao18, maze2018iarpa}. More recently, several have focused on balancing demographics in \gls{fr} data~\cite{hupont2019demogpairs, wang2018racial, karkkainen2019fairface, merler2019diversity}. Diversity in Faces (D\emph{i}F) came first~\cite{merler2019diversity}, which \tipr{came without} identity labels. D\emph{i}F is no longer available for download. Others released data with demographics balanced and omitted identity labels~\cite{wang2018racial, karkkainen2019fairface}. Hupont~\etal~\cite{hupont2019demogpairs} propose DemogPairs {balanced} across six subgroups of 600 identities from CASIA-WebFace (CASIA-W)~\cite{yi2014learning}, VGG~\cite{schroff2015facenet}, and VGG2~\cite{Cao18}. Our BFW includes eight subgroups (\ie split the African/Indian subgroup used in DemogPairs into separate groups, Black and Indian), 800 identities, and more face samples per identity. \tipr{We only use the VGG datasets, not CASIA-Web, to test a broader range of models} (\ie even models trained on CASIA-Web). {With public resources used to train existing models,} we built \gls{bfw} \tipr{using only} VGG2 to minimize conflicts in {the} overlap between train and test. Table~\ref{tab:bfw:attributes} compares our data with the others.

\begin{figure}[t!] 
\glsunset{fpr}
\centering
\centering
\includegraphics[trim={2mm 0 0 0},clip, width=.8\linewidth]{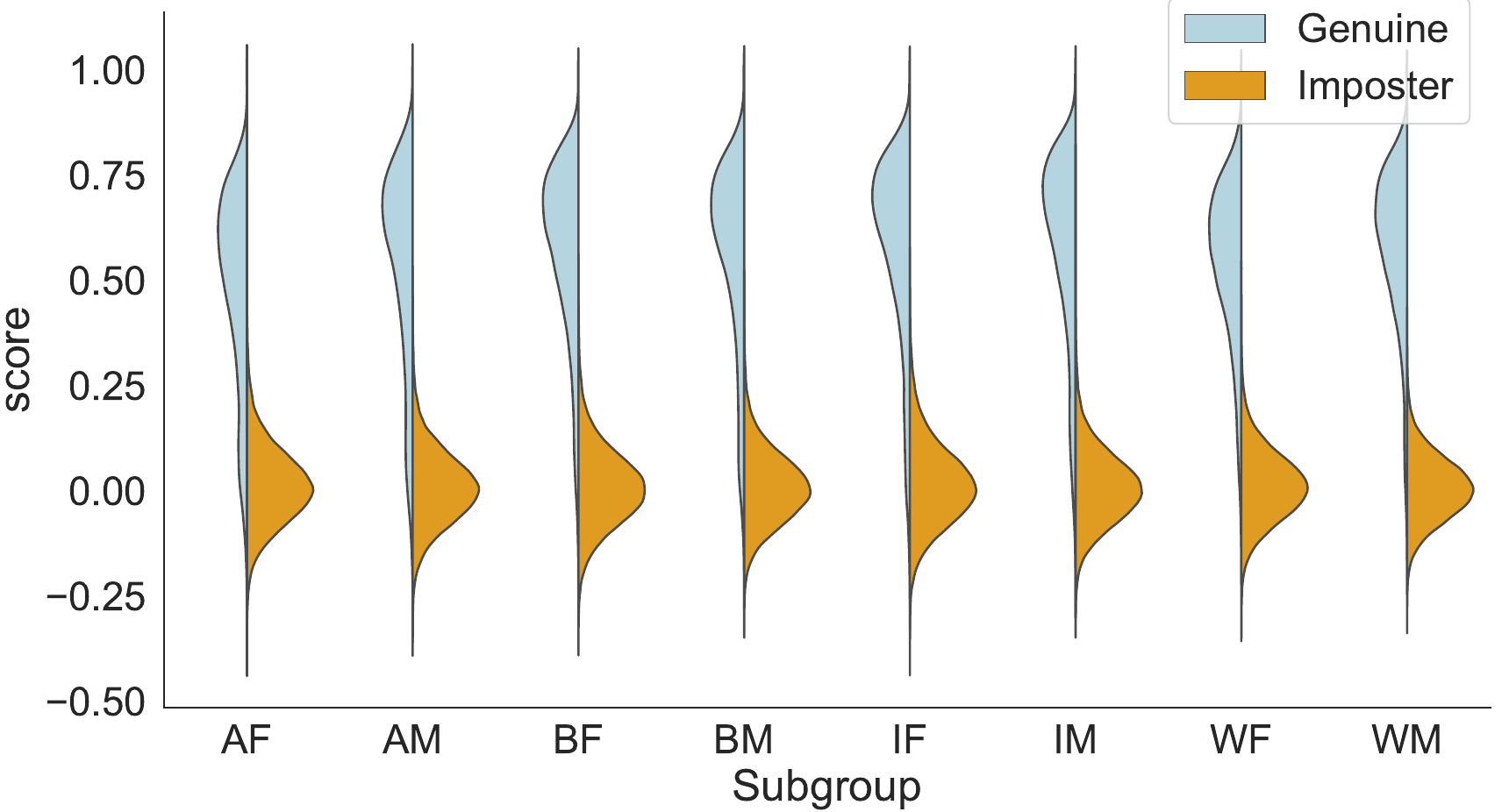}
	\caption{{\textbf{Score distributions per subgroup.} \emph{Imposters} have {$\mu\approx$0.0} but with variations in upper percentiles. \emph{Genuine} pairs vary in mean and spread. A threshold varying across subgroups yields a constant {\gls{fpr}.}}} \label{fig:detection-model} 
\end{figure} 

\tipr{Khan~\etal~\cite{zaid2021bias} study the limitations of face datasets with racial categories, including \gls{bfw}. The authors \tipr{note} the challenges of creating precise definitions of subgroups and measuring  the self-consistency of face datasets {and} cross-dataset consistency. (\gls{bfw} \tipr{is} approximately as consistent as other datasets.) Khan~\etal \tipr{also note} the challenges of racial types in science {and} the problems that \tipr{arise} without them (\eg generative models generating only Caucasian faces).}}

\glsreset{da}
\subsection{Feature alignment / Domain adaptation}
\Gls{da} employs labeled data from the source domain to \tipr{generalize well to the typically label-scarce target domain, which relieves the high costs and burden of labeling data by reducing the required amount~\cite{ding2018graph,peng2017visda,saito2019semi}.} {We can roughly classify DA} as a semi-supervised DA~\cite{saito2019semi,qin2022semi,qin2022robust} or an unsupervised one~\cite{shu2018dirt}, according to access to target labels. The crucial challenge toward DA is the distribution shift of features across domains (\ie domain gap), which violates the distribution-sharing assumption of conventional machine learning problems. {In our case, the domains are the subgroups} (Table~\ref{tab:bfw:counts}).

{Some} feature alignment (FA) methods \tipr{attempt} to project the raw data into a shared subspace where certain feature divergences or distances confuse groups. Many develop methods following this paradigm, such as correlation alignment~\cite{sun2015subspace}, maximum mean discrepancy~\cite{long2013transfer}, and geodesic flow kernel~\cite{gong2012geodesic,gopalan2011domain}. Adversarial domain alignment methods (\ie DANN~\cite{ganin2016domain}, ADDA~\cite{tzeng2017adversarial}) \tipr{design} a zero-sum game between a domain classifier (\ie discriminator) and a feature generator. The discriminator can not differentiate the source and target features if it \tipr{mixes the features of different domains.} More recently, learning well-clustered target features has proven helpful in conditional distribution alignment. DIRT-T~\cite{shu2018dirt} and MME~\cite{saito2019semi} use an entropy loss on target features to group them as multiple clusters in the feature space implicitly. This helps keep the discriminative structures through adaptation. By adjusting the sensitivities in true scores, we align the score distributions of the subgroups (Fig.~\ref{fig:detection-model}).

\glsreset{bfw}\glsunset{dp}

\subsection{Protecting demographic information in \gls{fr}}

    For reasons of privacy and protection, recent attempts \tipr{remove} demographic features from the raw face images~\cite{zhang2018mitigating, dhar2020adversarial, 8698605}. These works recognize the importance of {maintaining} identity information in facial features while ridding it of evidence of demographics. Our model inherently does this as part of the {target, aiming} for \tipr{the} inability to recognize subgroups. Some achieve this using adversarial learning on top of the features via a Minimax filter~\cite{hamm2017minimax}: maximizing the attributes loss while minimizing the target task. More recently, several \tipr{treat} it as a minimization problem by reversing the gradient of the protected classifier. For instance, Bertran~\etal~\cite{bertran2019adversarially} learn a projection that maps images to an embedding space {to disallow inference in} gender information. Similar to this, we also aim to rid the data of demographic knowledge. However, the difference is that we learn to protect demographics in the facial features often stored in place of imagery (\ie we map from a biased to a non-bias feature space). Ray~\etal~\cite{roy2019mitigating} follow the same path (\ie {map image-to-feature} with demographic information protected). Again, we aim to {preserve} a database of facial features, with no assumption that the initial model or raw images are accessible. 
    
    Wu~\etal~\cite{9207852, wu2018towards} present a method for removing gender information and preserving privacy in videos while maintaining sufficient information for action classification. They achieve this by learning a filter that selectively degrades the video. Their contribution differs from ours in that it operates in the image domain rather than the learned embedding domain. They also target action recognition rather than recognition applications.

\tipr{In fact,} several works aim to hide attribute information in image space. For instance, Othman~\etal~\cite{10.1007/978-3-319-16181-5_52} learn to morph faces to suppress gender {and preserve identity} information in the image space. Guo~\etal~\cite{GUO2019320} \tipr{map} the image to noise by encrypting the photo, such that the encoder decodes the identity without the ability to \tipr{recognize} gender. {Ma~\etal~\cite{ma2019lightweight} design protocols for transferring facial features via a cascade of classifiers in their lightweight {privacy-preserving} adaptive boosting (AdaBoost) framework.} Dhar~\etal~\cite{dhar2021pass} \tipr{attempts} to remove the attributes information from a pre-trained CNN with the help of discriminators and an injected generator layer. However, it is required to use multiple binary discriminators with correspondence with each attribute. Instead, we apply a single multi-class classifier to fulfill such an object, ensuring dense computation and improving model efficiency. 

\tipr{The proposed differs from previous work in the underlying data assumption: here, there is access to facial features, not the imagery, which is often the case in production. Images are processed once to reduce computation. Also, face features are compressed representations, making them much less expensive to store and transmit.}

\section{\gls{bfw}}\label{subsec:data} 
\begin{figure*}[t!]
    \centering
    \glsunset{af}
    \glsunset{am}
    \glsunset{bf}
    \glsunset{bm}
    \glsunset{if}
    \glsunset{im}
    \glsunset{wf}
    \glsunset{wm}
    \centering
    \begin{subfigure}[t]{.13\linewidth}
        \centering
        \includegraphics[width=\linewidth]{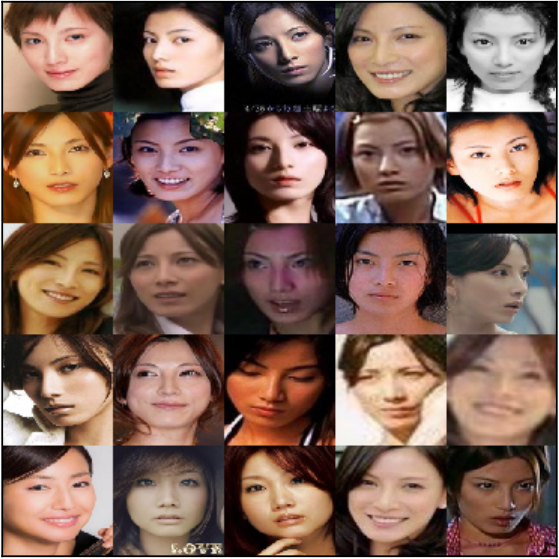}
        \caption{\gls{af}.}
        \label{subfig:af}
    \end{subfigure}%
    \hspace{1mm}
    \begin{subfigure}[t]{.13\linewidth}
        \centering
        \includegraphics[width=\linewidth]{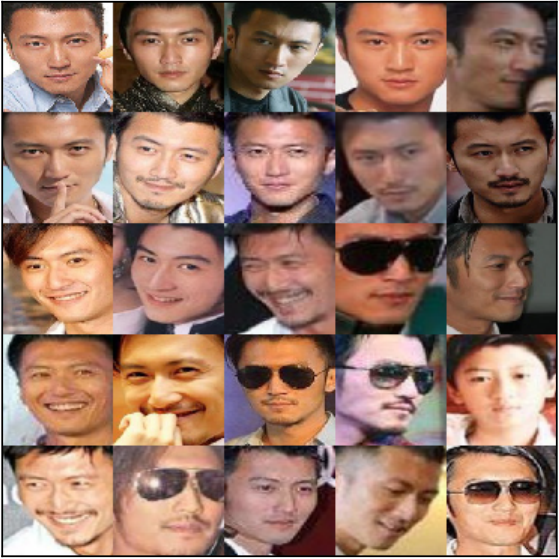}
        \caption{\gls{am}.}
         \label{subfig:am}
    \end{subfigure}
     \hspace{1mm}
    \begin{subfigure}[t]{.13\linewidth}
        \centering
        \includegraphics[width=\linewidth]{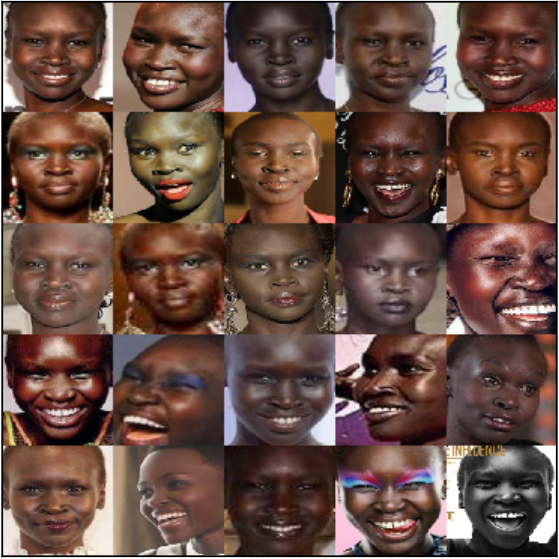}
        \caption{\gls{bf}.}
        \label{subfig:bf}
    \end{subfigure}%
     \hspace{1mm}
    \begin{subfigure}[t]{.13\linewidth}
        \centering
        \includegraphics[width=\linewidth]{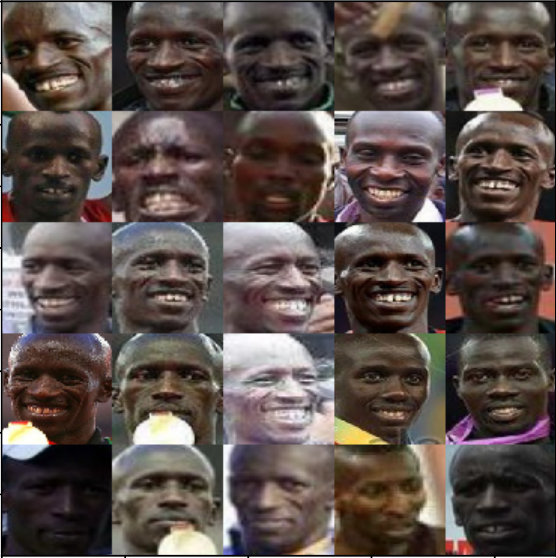}
        \caption{\gls{bm}.}
         \label{subfig:bm}
    \end{subfigure}\\
    \begin{subfigure}[t]{.13\linewidth}
        \centering
        \includegraphics[width=\linewidth]{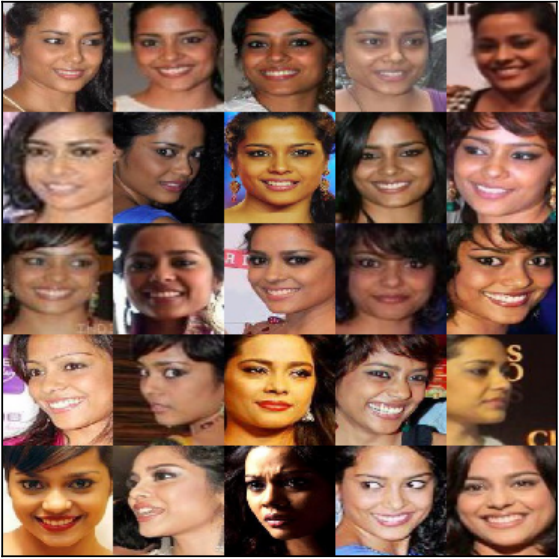}
        \caption{\gls{if}.}
        \label{subfig:if}
    \end{subfigure}%
     \hspace{1mm}
    \begin{subfigure}[t]{.13\linewidth}
        \centering
        \includegraphics[width=\linewidth]{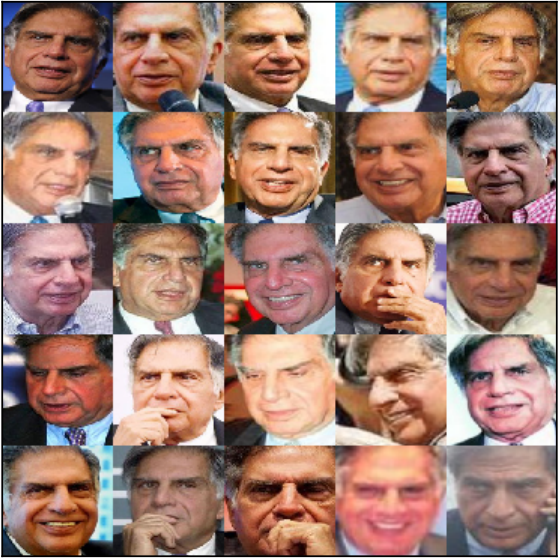}
        \caption{\gls{im}.}
         \label{subfig:ism}
    \end{subfigure}
     \hspace{1mm}
        \begin{subfigure}[t]{.13\linewidth}
        \centering
        \includegraphics[width=\linewidth]{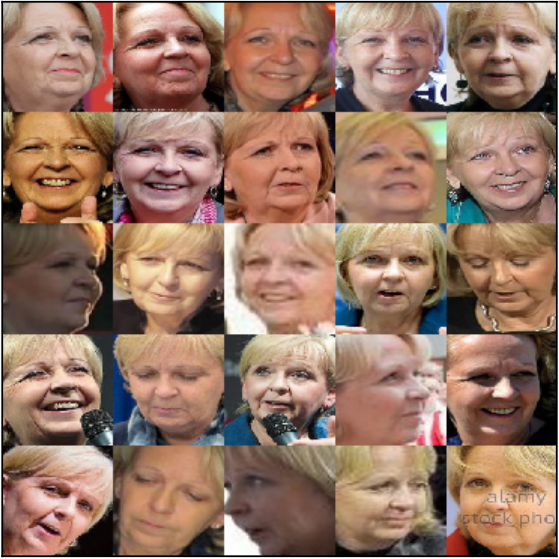}
        \caption{\gls{wf}.}
        \label{subfig:wf}
    \end{subfigure}%
     \hspace{1mm}
    \begin{subfigure}[t]{.13\linewidth}
        \centering
        \includegraphics[width=\linewidth]{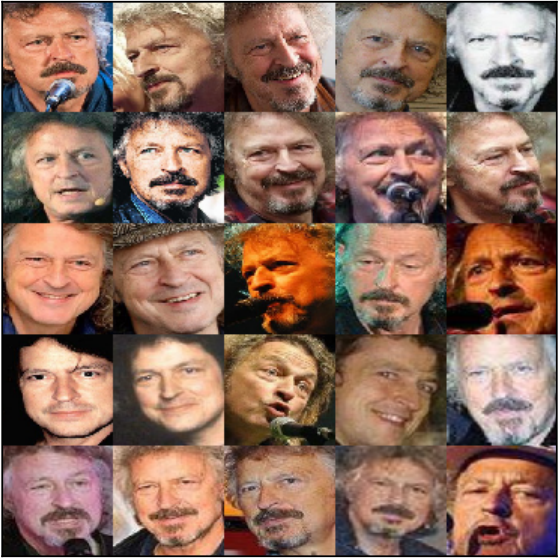}
        \caption{\gls{wm}.}
         \label{subfig:wm}
    \end{subfigure}\\
    
    \caption{\textbf{Samples of \gls{bfw}.} We show 25 samples for each subgroup for a randomly selected subject.}\label{fig:face-montage}
\end{figure*}

\gls{bfw} provides balanced data across ethnicity (\ie Asian (A), Black (B), Indian (I), and White (W)) and gender (\ie Female (F) and Male (M))--{eight} demographics referred to as subgroups (Fig.~\ref{fig:face-montage}). As in Table~\ref{tab:bfw:attributes}, \gls{bfw} has an equal number of subjects per subgroup (\ie 100 subjects per subgroup) and faces per subject (\ie 25 faces per subject). Note that the key difference between \gls{bfw} and \gls{dp} is in the additional attributes and the increase in labeled data; the differences {between} RFW and FairFace are in the identity labels and distributions (Table~\ref{tab:bfw:counts}).

\begin{table*}
\glsunset{rfw}
\glsunset{dp}
\vspace{5mm}
\centering
    \caption{\textbf{\gls{bfw} features compared to related resources.} Compared with \gls{dp}, \gls{bfw} provides more samples per subject and subgroup per set using only VGG2. \gls{rfw} supports a different task (\ie subgroup classification). {\gls{rfw}} and FairFace focus on {race distribution} without the support of identity labels.}\label{tab:bfw:attributes} 
    
        \begin{tabular}{rccccccc}
    \multicolumn{2}{c}{\cellcolor{red!20}\textbf{Database}} & \multicolumn{3}{c}{\cellcolor{blue!20}\textbf{Number of}}& \multicolumn{3}{c}{\cellcolor{green!20}\textbf{Balanced Labels}}\\
    \cellcolor{red!5}\textbf{Name} & \cellcolor{red!10}\textbf{Source} &\cellcolor{blue!5} \textbf{Faces} &  \cellcolor{blue!10}\textbf{IDs} & \cellcolor{blue!15}\textbf{Subgroups} & \cellcolor{green!5}\textbf{ID} & \cellcolor{green!10}\textbf{E} & \cellcolor{green!15}\textbf{G}\\
    \gls{rfw}~\cite{wang2018racial}     &  MS-Celeb-1M &$\approx$80,000&$\approx$12,000& 4 & \xmark & \checkc &\xmark \\
    \gls{dp}~\cite{hupont2019demogpairs}     & CASIA-W, VGG (+2) & 10,800& 600 & 6 &\checkc& \checkc &\checkc \\

    FairFace~\cite{karkkainen2019fairface} & Flickr, Twitter, Web & 108,000 & -- & 10 &\xmark& \checkc &\checkc\\

    \gls{bfw} (ours)~\cite{robinson2020face} & VGG2 & 20,000 & 800 &8 & \checkc & \checkc &\checkc 
    \end{tabular}
\glsreset{rfw}
\glsreset{dp}

\end{table*}

\begin{table*}
\centering
     \centering
     \caption{\textbf{Data statistics, vocabulary, and scores for subgroups as part of our \gls{bfw} data.} \textit{Top:} Specifications and subgroup definitions for BFW. \textit{Middle:} pair counts. \textit{Bottom:} accuracy using a global threshold $t_g$ and the optimal threshold $t_o$, and accuracy per subgroup. Columns are grouped by race and gender. Inconsistent ratings across subgroups. Subgroup acronyms used.}
    \begin{tabular}{r c c c c c c c c l}
        & \multicolumn{2}{c}{\cellcolor{red!20}\textbf{Asian (A)}} & \multicolumn{2}{c}{\cellcolor{blue!20}\textbf{Black (B)}}  & \multicolumn{2}{c}{\cellcolor{green!20}\textbf{Indian (I)}}& \multicolumn{2}{c}{\cellcolor{orange!20}\textbf{White (W)}}\\
         & \cellcolor{red!7}\textbf{Female (AF)} & \cellcolor{red!15}\textbf{Male (AM)} & \cellcolor{blue!7}\textbf{BF} & \cellcolor{blue!15}\textbf{BM}& \cellcolor{green!7}\textbf{IF} & \cellcolor{green!15}\textbf{IM} & \cellcolor{orange!7}\textbf{WF} & \cellcolor{orange!15}\textbf{WM}&\textbf{Aggregated}\\ 

    \rowcolor{lightgray!0}   \textbf{No. Faces}  &  2,500&  2,500& 2,500 &2,500 & 2,500 & 2,500 & 2,500 &2,500 &20,000 \\ 
       \rowcolor{lightgray!0}      \textbf{No. Subjects} & 100& 100& 100  & 100  & 100  & 100& 100 &100&800  \\ 
        \rowcolor{lightgray!0}     \textbf{No. Faces / subject}  & 25& 25    & 25 & 25 & 25  & 25  &  25 & 25 & 25\\ 
       \rowcolor{gray!8}         \textbf{No. Positive} &  30,000& 	30,000& 30,000 &30,000 & 30,000 &30,000&30,000 & 30,000 &240,000 \\ 
     \rowcolor{gray!8}        \textbf{No. Negative} &85,135&  85,232& 85,016  & 85,141 & 85,287  & 85,152& 85,223 &	85,193&681,379  \\ 
       \rowcolor{gray!8}  \textbf{Total} & 115,135 & 115,232 &115,016 &115,141 & 115,287 & 115,152 & 115,223& 115,193& 921,379\\ 

     \rowcolor{lightgray!0}        \textbf{Acc}$@\mathbf{t_g}$ & 0.876 & 0.944 &0.934 &0.942 &0.922&0.949 &0.916 &0.918&0.925$\pm$0.022 \\ 
    \rowcolor{lightgray!0}         $\mathbf{t_o}$ & 0.235 &  0.274 & 0.267&0.254  &0.299 & 0.295& 0.242 &0.222&0.261$\pm$0.025\\ 
     \rowcolor{lightgray!0}        \textbf{Acc}$\mathbf{@t_o}$ & 0.916 &0.964 &0.955 &0.971 &0.933 &0.958 & 0.969 &0.973 & 0.955 $\pm$ 0.018\\ 
    \end{tabular}
    \centering

\label{tab:bfw:counts} 
\end{table*}

{We built \gls{bfw}} with VGG2~\cite{Cao18} by using classifiers on the list of names and then the corresponding face data. Specifically, we ran a name-ethnicity classifier~\cite{ambekar2009name} to generate the initial list of subject proposals. Then, the corresponding faces with ethnicity~\cite{fu2014learning} and gender~\cite{levi2015age} classifiers further refined the list. Next, we manually validated, keeping only the genuine members of the respective subgroup. {We then limited faces for each subject to} 25 faces selected at random. Thus, \gls{bfw} {costs} minimal human input, having {generated the proposal lists} by automatic machinery.

\begin{figure}[t!]
  \centering
  \glsreset{ita}
\glsunset{fst}
    \includegraphics[width=.8\linewidth]{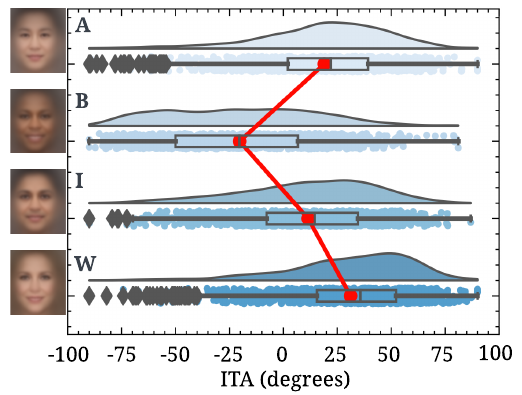}
   \caption{\textbf{\gls{ita} values by racial group.} Smaller \gls{ita} values are darker: Black (B) is the darkest, and Asian (A) and white (W) are the lightest, which is consistent with the facial color tones of the different subgroups. \tipr{The \gls{ita} for mean faces (left column) per race matches the mean ITA (red line).}}
    \label{fig:ita:violin}
    \glsreset{ita}
\glsreset{fst}
\end{figure}

{In summary, four experts in \gls{fr} manually validated all the data: first, the validation of individuals per subgroup was conducted (\ie inspect that all subjects belong to the assigned subgroups), and then {the} faces of the individual (\ie verify that each face instance belongs to the identity). {We only kept the subjects and samples verified as true by all annotators.} See our conference paper for additional details~\cite{robinson2020face}.}
    \begin{figure*}[t!]
        \centering
     
        \includegraphics[width=.85\linewidth]{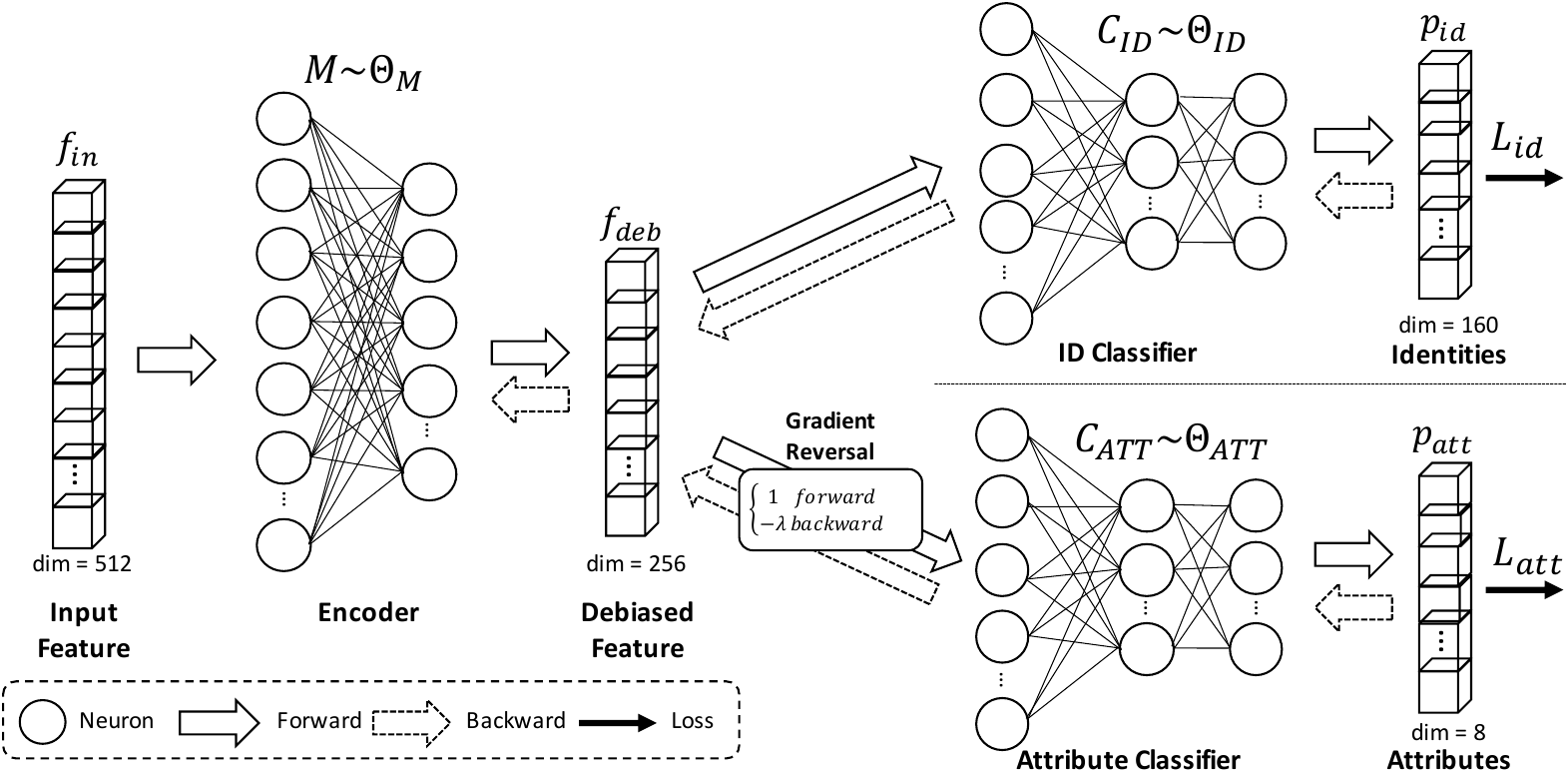}
        \caption{{\textbf{De-biasing framework.} The framework used to project facial {features} into a space that (1) preserves identity information (\ie $C_{ID}$) and (2) removes subgroup knowledge (\ie $C_{ATT}$). The \emph{gradient reversal}~\cite{ganin2015unsupervised} flips the sign of the error from $C_{ATT}$ to $M$ by a scalar $\lambda$ during training.}}\label{fig:framework}
    \end{figure*}
    
{We determined the subgroups of} \gls{bfw} based on physical features most common \tipr{among} the respective subgroup~\cite{robinson2020face}. {We can regard this} as multiple domains {because of} the feature {distribution} mismatch across these subgroups. However, the assumption {is} that a discrete label that can describe an individual is {imprecise.} {The} assumption allows for a finer-grain analysis of {the} subgroup and is a step in the right direction. Thus, we refute any claim that our efforts here are the {ultimate} solution. The data and proposed machinery are merely an attempt to establish a foundation for future work to extend. {The two genders} for the four ethnic groups make up the eight subgroups of the \gls{bfw} dataset (Fig.~\ref{fig:face-montage}). Formally put, the tasks addressed have labels for gender $l^g\in\{F, M\}$ and ethnicity $l^e\in\{A, B, I, W\},$ where the $K$ subgroups (\ie demographics) are then $K=|l_g|*|l_e|=8$.

{
\subsection{The data subgroups}
K$\ddot{a}$rkk$\ddot{a}$inen~\etal~\cite{karkkainen2019fairface} note that physical attributes correlate with the human race, while ethnicity is culturally based. Still, people often use race and ethnicity interchangeably. We refer to the U.S. Census Bureau to choose subgroups. Such labels are oversimplified~\cite{yudell2016taking} and are not precisely defined \cite{zaid2021bias}. From these limitations, the categories can show value for sub-group analysis of bias in computer vision~\cite{robinson2020face}.

Dermatologists diagnose sun exposure risks by manual inspection of the tone of a subject's skin with a label called the  \gls{fst}. Because of the need for manual {review} by multiple experts can be challenging to collect such data. Merler~\etal~\cite{hill1973diversity} propose a digital image processing scheme to characterize the skin tones of faces in their dataset, \gls{dif}. The authors reference an earlier study that revealed a correlation between the \gls{mi}, a measure objectively measured by reflectance spectrophotometry~\cite{10.1001/jamadermatol.2013.6101}, and the \gls{ita}: a practical measure to categorize skin tones, as the \gls{fst} can be determined digitally. \Gls{ita} is calculated from pixels in CIE-Lab color space as follows:

\begin{equation}\label{eq:ita}
\text{ITA}=\arctan{\left(\frac{L-50}{b}\right)}\times\frac{180^{\circ}}{\pi},
\end{equation}

 \noindent where larger lightness $L$ and smaller blue-yellow $b$ yield larger \glspl{ita}.  We proceed following~\cite{hill1973diversity} (and perhaps less like Kinyanjui~\etal~\cite{DBLP:journals/corr/abs-1910-13268}). We mask out target regions of the skin. However, instead of splitting face into {areas} based on detected landmarks, we segment faces, masking out all but the flat {areas} to avoid shadowing (\ie omitting the nose, eyes, mouth, and hair).\footnote{\href{https://github.com/shaoanlu/face_toolbox_keras}{https://github.com/shaoanlu/face\_toolbox\_keras}} With the pixels transformed from RGB to CIE-Lab, we removed $L$ and $b$ values more than one standard deviation of the respective mean for that face. we further mitigated concerns of outlier pixels by smoothing {them} out via a mean filter. Finally, the mean of pixel-wise \gls{ita} values {is} calculated for a face (Eq.~\ref{eq:ita}).

Fig.~\ref{fig:ita:violin} shows the distribution of all \gls{ita} values. The values line up within expectation: the smaller the \gls{ita} (in degrees), the darker the skin tone. Notice the left tail of the Black, Indian, Asian, and White go from the \tipr{densest to the scarcest.} {Furthermore, the mean (\ie vertical line in the figure) shifts right for the lighter-toned skin subgroups. There is a significant spread in values within racial groups, partly due to the varied lighting conditions.}}

\glsreset{fv}\glsreset{fr}
\section{Methodology}\label{sec:proposed}

\tipr{We first introduce the bias and privacy concerns of \gls{fv} systems, and then we explain our method for addressing these issues.} Specifically, we review the problem statement, the \gls{bfw} dataset, and the proposed framework.

    \subsection{Problem statement}
        \gls{fv} systems \tipr{infer} the \tipr{likelihood} that a pair of faces share the same identity. Verification is often solved like traditional \gls{fr}. Specifically, a model is trained on a set of identities and then used to encode faces (\ie embed faces). The closeness of the resulting vectors is a single score; typically, cosine similarity is used~\cite{nguyen2010cosine}. \tipr{The goal is to learn the optimal score that separates valid from false pairs.} The threshold is the decision boundary in score space, \ie the \emph{matching function}. \tipr{As demonstrated in our previous work, the optimal threshold changes between subgroups~\cite{robinson2020face}. Our prior solution was to learn a threshold per subgroup, which assumes the subgroup is known. We now aim to project features to a space that simultaneously preserves the identity and removes evidence of the subgroups. As we show, the results are less biased, while demographic privacy is preserved.}

    \subsubsection{The matching function}\label{subsec:metrics}
        A real-valued similarity score $\mathrm{R}$ assumes a discrete label of $Y=1$ for \emph{genuine} pairs (\ie a true match) or $Y=0$ for an \emph{imposter} (\ie untrue match). {We map the actual real number} to a discrete label by $\hat{Y}=\mathbbm{I}\{\mathrm{R}>\theta\}$ for some {pre-defined} threshold $\theta.$ {We can express the aforementioned} as \emph{matcher} $\mathrel{d}$ operating as

        \begin{equation}\label{eg:matcher}
            f_{boolean}(\vec{x}_i, \vec{x}_j) = \mathrel{d}(\vec{x}_i, \vec{x}_j) > \theta,
        \end{equation}

   \noindent{where} the face {features} in $\vec{x}$ being the $i^{th}$ and $j^{th}$ sample - a conventional scheme in the \ac{fr} research communities~\cite{LFWTech}. We use cosine similarity as the \emph{matcher} in Eq.~\ref{eg:matcher}, which produces a score in closeness for the $i^{th}$ and $j^{th}$ faces (\ie {$l^{th}$} face pair) by
    $d(\vec{x}_i, \vec{x}_j)=s_l= \frac{f_i\cdot f_j}{||f_i||_2||f_j||_2}.$ The decision boundary formed by {the} threshold $\theta$ controls the level of \emph{acceptance} and \emph{rejection}. Thus, $\theta$ inherits a trade-off between {sensitivity} and specificity. {The value of $\theta$ depends on the purpose of the system. For instance, in \tipr{security,} there is a need for higher sensitivity (\ie smaller $\theta$).}
    Specifically, {the trade-off involves \gls{fnr} that attempts to pass but falsely rejects—a Type 1 Error.} {Mathematically, it relates by
    $$
    \text{FNR}=\frac{\text{FN}}{\text{P}}=\frac{\text{FN}}{\text{FN}+\text{TP}}=1-\text{TPR} = 1 - \frac{\text{TP}}{\text{FN}+\text{TP}},
    $$\noindent
    with positive counts $P$. 
    
    The other error type contributes to the \gls{fpr}, {the Type II Error, which is when an \textit{imposter} falsely passes:}
    
    $$
    \text{FPR}=\frac{\text{FP}}{\text{N}}=\frac{\text{FP}}{\text{FP}+\text{TN}}=1-\text{TNR}=1-\frac{\text{TN}}{\text{FP}+\text{TN}},
    $$
    \noindent
    where {the number of} negatives is $N,$ {with metrics} \gls{tn}, \gls{fp}, \gls{tnr}, and \gls{fpr}.} 
    The geometric relationships of the metrics related to the score distributions and the choice {of} threshold {show} the trade-offs in error rates (Fig.~\ref{fig:metrics}).
    
   \tipr{The parameter} $\theta$ determines the error rate on held-out validation, specific to the {use case.} {Researchers tend to set it} {for top} \tipr{performance, while others analyze $\theta$ as a range of values to generate plots and assess the trade-offs.} The held-out validation and test sets share data distributions as a single source partitioned into subsets (\ie train, validation, test). {We transfer} the decision boundary in score space, which maximizes the performance to the pin-point (\ie 1D) decision boundary—the {floating-point} value {spans} [0, 1].

\begin{figure*}[t!h]
\glsreset{fpr}
\begin{subfigure}[!ht]{.34\linewidth}
    \includegraphics[width=\linewidth]{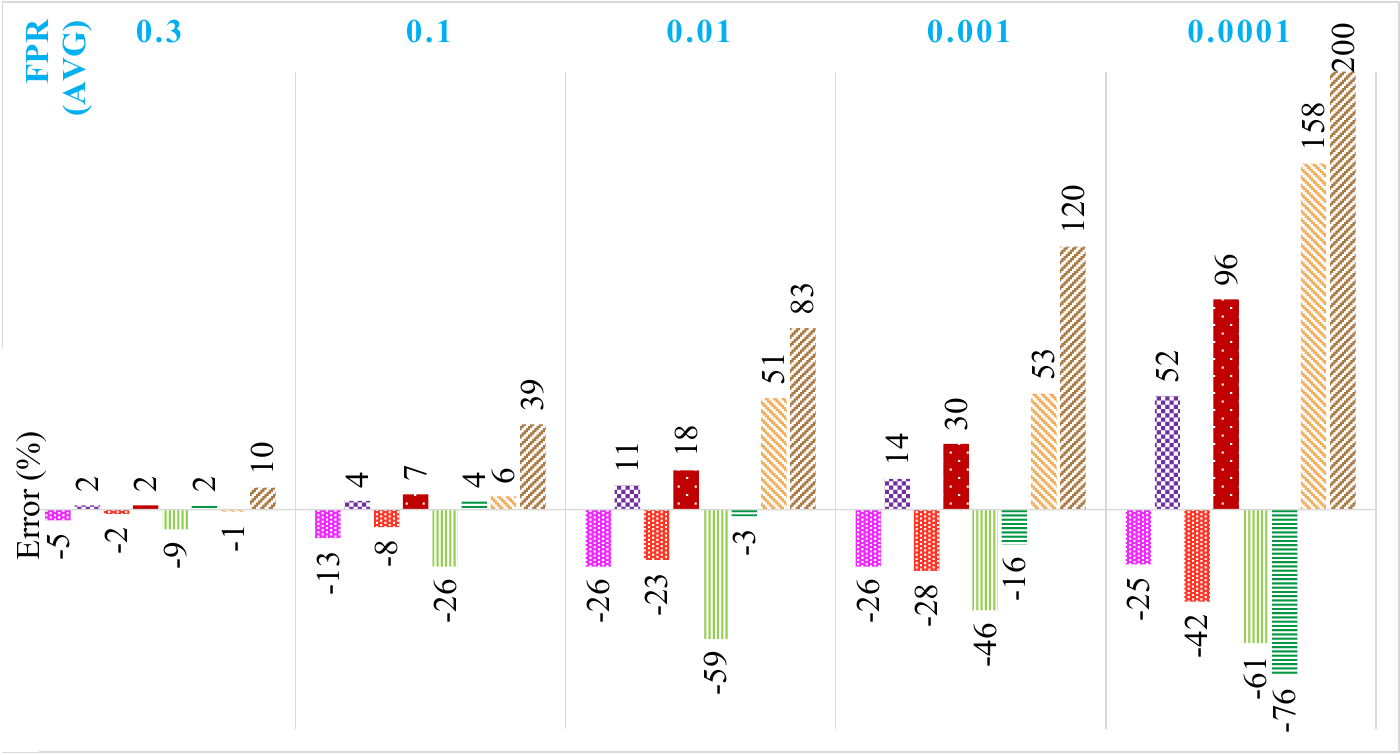}
    \caption{Global}
\end{subfigure}
\begin{subfigure}[h!t]{.31\linewidth}
    \includegraphics[width=\linewidth]{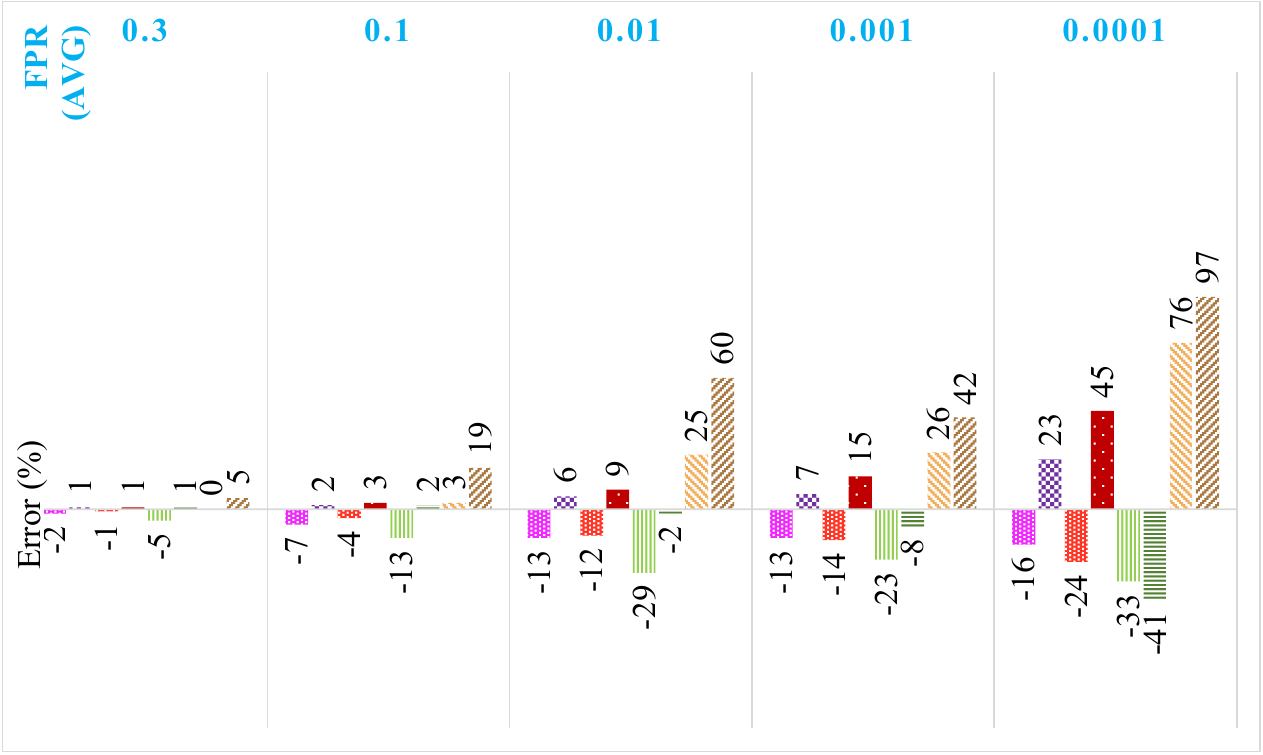}
    \caption{Proposed}
    \label{pdf/1bbb.pdf}
\end{subfigure}
\begin{subfigure}[h!t]{.3\linewidth}
    \includegraphics[width=\linewidth]{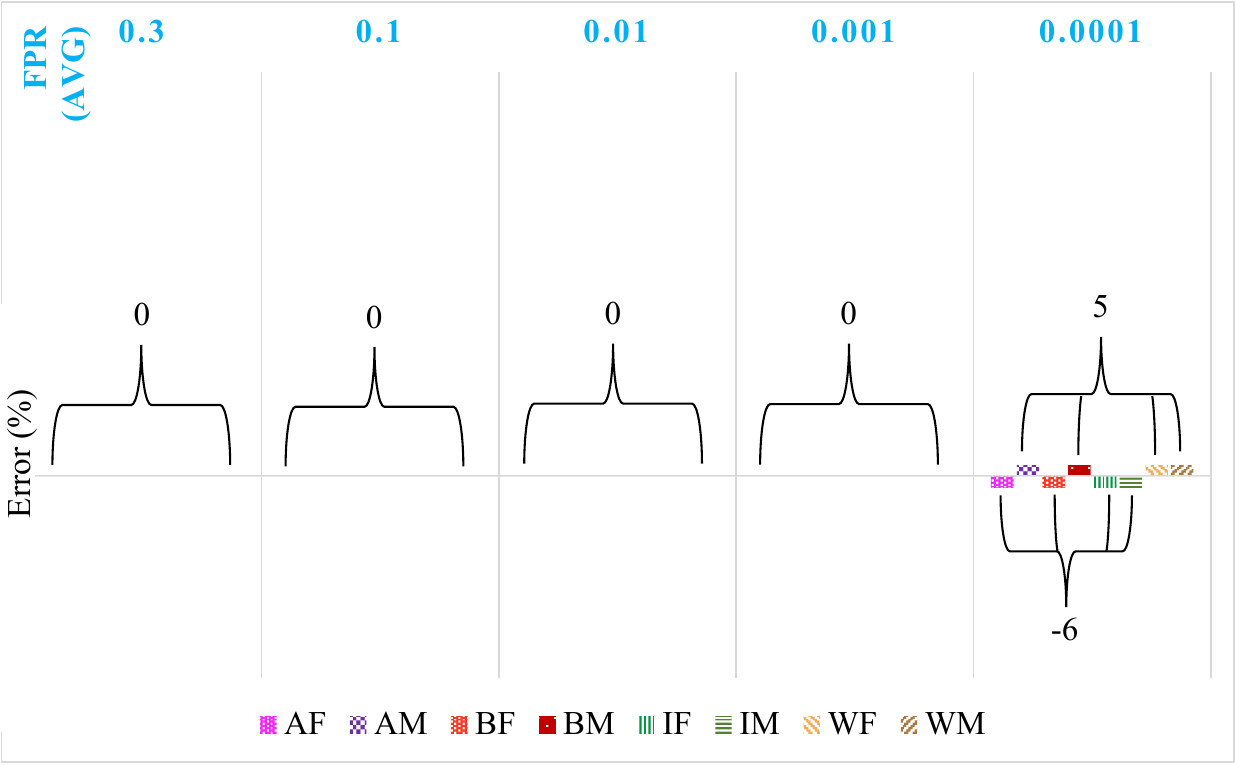}
    \caption{Subgroup-Specific}
\end{subfigure}
    \caption{\textbf{Percent difference in the \gls{fpr} from the mean.} \emph{Global} threshold ($t_g$) \tipr{differs as much as 200\% (\ie WM at \gls{fpr}=0.0001);} the \tipr{female subgroups mostly perform below average; the males are mostly above (\textcolor{red}{a}).} \emph{Subgroup-specific} thresholds ($t_o$) reduce the percent difference close to zero (\textcolor{red}{c}). The proposed method does not assume knowledge of attributes like $t_o,$ reduces the discrepancies of subgroups (\textcolor{red}{b}). Similar to Fig.~\ref{tab:ethnicy-far}, the variations are nearly halved from the baseline using the proposed.}
     \label{fig:page1-teaser-barplot}
\end{figure*}

\subsubsection{Feature alignment}
{The} tuple $\mathcal{D}=\{(\mathbf{x}_i,y_i) \in \mathcal{X} \times \mathcal{Y} \}_{i=1}^{N}$ represents domain $\mathcal{D},$ with $\mathcal{X}$ and $\mathcal{Y}$ representing the input feature space and output label space, respectively. {\gls{fr} algorithms aim} to learn a mapping function (\ie {a} hypothesis): ${\eta} : \mathcal{X} \rightarrow \mathcal{Y},$ {assigning vectors} with a semantic identity label.

Mathematically, {we denote} the labeled source domain $\mathcal{D}_S$ and the unlabeled target domain $\mathcal{D}_T$ as  $\mathcal{D}_S = \{(\mathbf{x}_i^s, y_i^s) \in \mathcal{X}_S \times \mathcal{Y}_S\}^{N_S}_{i=1}$ and  $\mathcal{D}_T = \{\mathbf{x}_i^t \in \mathcal{X}_T \}^{N_T}_{i=1}$ with the sample count $N_S=|\mathcal{D}_S|$ and $N_T=|\mathcal{D}_T|$ corresponding to the $i$-th sample (\ie $\mathbf{x}_i \in \mathbb{R}^{d}$) and label (\ie $y_i \in \{1,..., K\}$). We further define $\mathcal{D}_S$ and $\mathcal{D}_T$ as tasks $\mathcal{T}_S$ and $\mathcal{T}_T,$ \tipr{respectively,} which {show} the exact label type(s) and the specific $K$ classes of interest. The goal is to learn an objective ${\eta}_S : \mathcal{X}_S \rightarrow \mathcal{Y}_S,$ then transfer to target $\mathcal{D}_T$ for $\mathcal{T}_T.$ By this, we leverage {knowledge} from both $\mathcal{D}_S$ for $\mathcal{D}_T$ {to get} $\eta_T.$ Since either domain has different marginal distributions (\ie $p({\mathbf{x}}^s)\not=p({\mathbf{x}}^t)$) and distinct conditional distributions (\ie {$p(y^t|{\mathbf{x}}^s)\not=p(y^t|{\mathbf{x}}^t),$)} a model trained on the labeled source usually performs poorly on the unlabeled target. A standard solution {to} {a} domain gap is to learn a model $f$ that aligns the features in a shared subspace by $p({f(\mathbf{x}}^s))\approx p({f(\mathbf{x}}^t)).$ 

\subsection{Proposed framework}

{We used both identity and subgroup labels} for the two objectives of the proposed framework (Fig.~\ref{fig:framework}). Specifically, $\mathcal{D} = \{\mathbf{x}_i, y_i^{id}, y_i^{att} \}^{N}_{i=1},$ where $\mathbf{x} \in \mathbb{R}^{d},$ $y^{id} \in \{1,...,I\}$ and $y^{att} \in \{1,...,K\}.$ Hence, we aim to learn a mapping $\mathbf{f}_{deb} = M(\mathbf{x}, \Theta_M)$ to a {lower-dimensional} space $\mathbf{f}_{deb} \in \mathbb{R}^{d/2}$ that preserves {the} identity information of the target via the identity loss $\mathcal{L}_{ID}.$ {Then, we} learn to do so without subgroup information, which we call the attribute (or subgroup) loss $\mathcal{L}_{ATT}.$ The total loss - the final objective $\mathcal{L} = \mathcal{L}_{ID} + \mathcal{L}_{ATT}$) - {is the sum of all losses.}

\begin{align}
\mathcal{L}_{ID} = - &\frac{1}{N} \sum_{i=1}^{N}  \sum_{k=1}^{I}  \mathbf{1}_{[k=y_i^{id}]}{\mathrm{\log}{({p}({y}=y_i^{id}| \mathbf{x}_i)})} ,\label{e2} \\
\mathcal{L}_{ATT}= - &\frac{1}{N} \sum_{i=1}^{N}  \sum_{k=1}^{K}  \mathbf{1}_{[k=y_i^{att}]}{\mathrm{\log}{( {p}({y}=y_i^{att}| \mathbf{x}_i})} , \label{e3}
\end{align}

\noindent{where} ${p}({y}=y_i^{id}| \mathbf{x}_i)$ and ${p}({y}=y_i^{att}| \mathbf{x}_i)$ represent \tipr{the probability conditioned on the identity and attribute, respectively.}

We added $\mathcal{L}_{ATT}$ to de-bias the features to remove variation in scores previously handled with a variable threshold.
Furthermore, a byproduct {is} these features that preserve identity information without knowledge of subgroups -- a critical concern in the privacy and protection of biometric data.

There are three groups of parameters (\ie $\Theta_M,$ {$\Theta_{ID},$} and $\Theta_{ATT}$) optimized by the objective (Fig.~\ref{fig:framework}). {Both classifiers, the identity $C_{ID}$ and the attribute $C_{ATT},$} are used to find a feature space that remains accurate to identity and not for subgroup by minimizing the empirical risk of $\mathcal{L}_{ID}$ and $\mathcal{L}_{ATT}$:

\begin{align}
    {\Theta}_{ID}^{*} =&\mathop {\arg \min} \limits_{{\Theta}_{ID}}  \mathcal{L}_{ID},\label{e4}\\
    {\Theta}_{ATT}^{*} =&\mathop {\arg \min} \limits_{{\Theta}_{ATT}}  \mathcal{L}_{ATT}.\label{e5}
\end{align}

Thus, a gradient reversal layer~\cite{ganin2015unsupervised} that acts as the identity during the forward pass while inverting the sign of the gradient {back-propagated} with a scalar $\lambda$ as the adversarial loss during training:

\begin{equation}
{\Theta}_{M}^{*} =\mathop {\arg \min} \limits_{{\Theta}_{M}} - \lambda \mathcal{L}_{ATT} +  \mathcal{L}_{ID}. \label{e6}
\end{equation}

Although the proposed learning scheme is simple, it proved effective for both objectives we seek to solve. Next, we illustrate the effectiveness of the results and provide an analysis.

\begin{figure}[!ht]
\centering
\begin{subfigure}[!ht]{\linewidth}
\centering
\resizebox{.8\textwidth}{!}{%
\begin{tabular}{l  c c c c c l }\toprule
        \textbf{\gls{fpr}}& 0.3 &0.1 & 0.01 & 0.001 & 0.0001&\\\midrule
     \multirow{3}{3mm}{\textbf{\gls{af}}}  &\cellcolor{blue!20} 0.990 & \cellcolor{blue!20} 0.867 & \cellcolor{blue!20} 0.516 & \cellcolor{blue!20} 0.470 &\cellcolor{blue!20}  0.465&\cellcolor{blue!20}\emph{G}\\[-4pt]
      &\cellcolor{blue!10} 0.996 & \cellcolor{blue!10}0.874 & \cellcolor{blue!10}0.521 &\cellcolor{blue!10} 0.475 & \cellcolor{blue!10}0.470&\cellcolor{blue!10}\emph{P-P}\\[-4pt]
      &\cellcolor{blue!5}1.000 & \cellcolor{blue!5} 0.882 & \cellcolor{blue!5} 0.524 & \cellcolor{blue!5} 0.478 & \cellcolor{blue!5} 0.474&\cellcolor{blue!5}\emph{S-S}\\[-1pt]
    \multirow{3}{3mm}{\textbf{\gls{am}}} &\cellcolor{blue!20} 0.994 & \cellcolor{blue!20} 0.883 & \cellcolor{blue!20} 0.529 &\cellcolor{blue!20}  0.482 &\cellcolor{blue!20}  0.477&\cellcolor{blue!20} \\[-4pt]
        &\cellcolor{blue!10}0.996 &\cellcolor{blue!10} 0.886 &\cellcolor{blue!10} 0.531 & \cellcolor{blue!10}0.484 &\cellcolor{blue!10} 0.479&\cellcolor{blue!10}\\[-4pt]
        &\cellcolor{blue!5}1.000 & \cellcolor{blue!5} 0.890 & \cellcolor{blue!5} 0.533 & \cellcolor{blue!5} 0.486 & \cellcolor{blue!5} 0.482&\cellcolor{blue!5}\\[-1pt]
    \multirow{3}{3mm}{\textbf{\gls{bf}}} &\cellcolor{blue!20} 0.991 &\cellcolor{blue!20}  0.870 &\cellcolor{blue!20}  0.524 &\cellcolor{blue!20}  0.479 &\cellcolor{blue!20}  0.473&\cellcolor{blue!20} \\[-4pt]
        &\cellcolor{blue!10}0.995 & \cellcolor{blue!10}0.875 &\cellcolor{blue!10} 0.527 & \cellcolor{blue!10}0.481 &\cellcolor{blue!10} 0.476&\cellcolor{blue!10}\\[-4pt]
        &\cellcolor{blue!5}1.000 & \cellcolor{blue!5} 0.879 & \cellcolor{blue!5} 0.530 & \cellcolor{blue!5} 0.484 & \cellcolor{blue!5} 0.480&\cellcolor{blue!5}\\[-1pt]
    \multirow{3}{3mm}{\textbf{\gls{bm}}} &\cellcolor{blue!20} 0.992 & \cellcolor{blue!20} 0.881 &\cellcolor{blue!20}  0.526 & \cellcolor{blue!20} 0.480 & \cellcolor{blue!20} 0.474&\cellcolor{blue!20} \\[-4pt]
        &\cellcolor{blue!10}0.995 &\cellcolor{blue!10} 0.886 &\cellcolor{blue!10} 0.529 & \cellcolor{blue!10}0.483 &\cellcolor{blue!10} 0.478&\cellcolor{blue!10}\\[-4pt]
        &\cellcolor{blue!5}1.000 & \cellcolor{blue!5} 0.891 & \cellcolor{blue!5} 0.532 & \cellcolor{blue!5} 0.485 & \cellcolor{blue!5} 0.480&\cellcolor{blue!5}\\[-1pt]
    \multirow{3}{3mm}{\textbf{\gls{if}}} &\cellcolor{blue!20} 0.996 &\cellcolor{blue!20}  0.881 &\cellcolor{blue!20}  0.532 & \cellcolor{blue!20} 0.486 &\cellcolor{blue!20}  0.481&\cellcolor{blue!20} \\[-4pt]
        &\cellcolor{blue!10}0.998 &\cellcolor{blue!10} 0.883 & \cellcolor{blue!10}0.533 &\cellcolor{blue!10} 0.487 &\cellcolor{blue!10} 0.483&\cellcolor{blue!10}\\[-4pt]
        &\cellcolor{blue!5}1.000 & \cellcolor{blue!5} 0.884 & \cellcolor{blue!5} 0.534 & \cellcolor{blue!5} 0.488 & \cellcolor{blue!5} 0.484&\cellcolor{blue!5}\\[-1pt]
    \multirow{3}{3mm}{\textbf{\gls{im}}} &\cellcolor{blue!20} 0.997 & \cellcolor{blue!20} 0.895 & \cellcolor{blue!20} 0.533 &\cellcolor{blue!20}  0.485 & \cellcolor{blue!20} 0.479&\cellcolor{blue!20} \\[-4pt]
        &\cellcolor{blue!10}0.998 & \cellcolor{blue!10}0.897 &\cellcolor{blue!10} 0.534 &\cellcolor{blue!10} 0.486 &\cellcolor{blue!10} 0.480&\cellcolor{blue!10}\\[-4pt]
        &\cellcolor{blue!5}1.000 & \cellcolor{blue!5} 0.898 & \cellcolor{blue!5} 0.535 & \cellcolor{blue!5} 0.486 & \cellcolor{blue!5} 0.481&\cellcolor{blue!5}\\[-1pt]
    \multirow{3}{3mm}{\textbf{\gls{wf}}} &\cellcolor{blue!20} 0.988 & \cellcolor{blue!20} 0.878 &\cellcolor{blue!20}  0.517 & \cellcolor{blue!20} 0.469 & \cellcolor{blue!20} 0.464&\cellcolor{blue!20} \\[-4pt]
        &\cellcolor{blue!10}0.992 &\cellcolor{blue!10} 0.884 & \cellcolor{blue!10}0.522 & \cellcolor{blue!10}0.472 &\cellcolor{blue!10} 0.468&\cellcolor{blue!10}\\[-4pt]
        &\cellcolor{blue!5}1.000 & \cellcolor{blue!5} 0.894 & \cellcolor{blue!5} 0.526 & \cellcolor{blue!5} 0.478 & \cellcolor{blue!5} 0.474&\cellcolor{blue!5}\\[-1pt]
    \multirow{3}{3mm}{\textbf{\gls{wm}}} &\cellcolor{blue!20} 0.989 & \cellcolor{blue!20} 0.896 &\cellcolor{blue!20}  0.527 &\cellcolor{blue!20}  0.476 &\cellcolor{blue!20}  0.470&\cellcolor{blue!20} \\[-4pt]
        &\cellcolor{blue!10}0.996 & \cellcolor{blue!10}0.901 & \cellcolor{blue!10}0.530 &\cellcolor{blue!10} 0.479 & \cellcolor{blue!10}0.474&\cellcolor{blue!10}\\[-4pt]
      &\cellcolor{blue!5}1.000 & \cellcolor{blue!5} 0.910 & \cellcolor{blue!5} 0.535 & \cellcolor{blue!5} 0.483 & \cellcolor{blue!5} 0.478&\cellcolor{blue!5}\\[-2pt]\midrule
    \multirow{3}{3mm}{\textbf{Avg.}} &\cellcolor{blue!20}0.992 & \cellcolor{blue!20} 0.881 & \cellcolor{blue!20} 0.526 & \cellcolor{blue!20} 0.478 &\cellcolor{blue!20}  0.473&\cellcolor{blue!20} \\[-4pt]
        &\cellcolor{blue!10}0.998 & \cellcolor{blue!10}0.886 & \cellcolor{blue!10}0.528 &\cellcolor{blue!10} 0.481 & \cellcolor{blue!10}0.476&\cellcolor{blue!10} \\[-4pt]
        &\cellcolor{blue!5}1.000 & \cellcolor{blue!5} 0.891 & \cellcolor{blue!5} 0.531 & \cellcolor{blue!5} 0.483 & \cellcolor{blue!5} 0.479&\cellcolor{blue!5} \\\bottomrule
\end{tabular}
}

\glsunset{tpr}
 \caption{\gls{tpr}@\gls{fpr} per subgroup.}\label{fig:far:a}
\end{subfigure}\\

\centering
\begin{subfigure}[t]{\linewidth}
\centering
\resizebox{.55\columnwidth}{!}{%
\begin{tikzpicture}

  \begin{axis}
    [
    ytick={1,2,3},
    yticklabels={S-S, P-P, G},
    ]
    \addplot+[color=blue,
    boxplot prepared={
      median=100.0,
      upper quartile=100.0,
      lower quartile=100.0,
      upper whisker=100.0,
      lower whisker=100.0
    },
    ] coordinates {};

    \addplot+[color=blue,
    boxplot prepared={
      median=99.6,
      upper quartile=99.7,
      lower quartile=99.5,
      upper whisker=99.8,
      lower whisker=99.3
    },
    ] 
    coordinates {};
        \addplot+[color=blue,
    boxplot prepared={
      median=99.2,
      upper quartile=99.5,
      lower quartile=99.0,
      upper whisker=99.7,
      lower whisker=98.7
    },
    ] coordinates {};

  \end{axis}
\end{tikzpicture}
}
    \caption{\gls{tpr}@\gls{fpr}=0.3.}
     \label{fig:far:b}
\end{subfigure}
    \caption{\textbf{\Gls{tpr} at a {\gls{fpr}.}} The last column of AF shows how the \gls{tpr} scores for the global (G) threshold, \emph{privacy-preserving} (P-P) features (\ie proposed), and \emph{subgroup-specific} (S-S) threshold (\ie baseline) go from darkest to lightest (labeled in the last column of AF). Higher is better (\ref{fig:far:a}). \tip{The spread of  \emph{G} scores across subgroups is larger than that of  \emph{S-S} scores, as shown clearly in (\ref{fig:far:b}),  which visualizes the left column in (\ref{fig:far:a}).} }\label{tab:ethnicy-far}
\glsunset{tpr}
\glsunset{fpr}
\end{figure}
\section{Experiments}\label{sec:experimental}

{We include two sets of experiments to show} the effectiveness of the proposed using our balanced \gls{bfw}~\cite{robinson2020face}. First, we evaluate verification performance. Specifically, we compare the \emph{global}, \emph{subgroup-specific}, and baseline.
Then, for the {privacy-preserving} claim, we compare the performance of models trained on top of de-biased features $f_{deb}$ with those of the original features $f_{in}.$ We present the problem statement, metrics and settings, and analysis for each. \tipr{An ablation study shows the performance \tipr{of} LFW~\cite{LFWTech}.}

\subsection{Common settings}
{We use Arcface (\ie ResNet-34) as the baseline (\ie $f_{in}$)~\cite{wang2018additive}.}
MS1M~\cite{guo2016ms} was the train set, {with} about 5.8 million faces for 85,000 subjects. We prepared the faces using MTCNN~\cite{zhang2016joint} to detect five facial landmarks. {We then applied a similarity transformation} to align the face by the five detected landmarks, from which we cropped and resized each to 96$\times$112. The RGB (\ie pixel values of [0, 255]) were normalized by centering about 0 (\ie subtracting 127.5) and then standardizing (\ie dividing by 128); {features} were later L2 normalized~\cite{wang2017normface}. The batch size was 200, and an {SGD} optimizer with a momentum {of} 0.9, weight decay 5e-4, and the learning rate started at 0.1 and {factored} by 10 two times when the error leveled. {We chose these settings} based on Arcface being among the best-performing \gls{fr} deep models. \emph{Off-the-shelf} \glspl{cnn} are typical solutions implemented in systems using \gls{fr} technology in research and practice. 

{We} used our \gls{bfw} dataset (Section~\ref{subsec:data}): the de-bias and privacy-based experiments use the {pre-defined} five-folds;
the ablation study on LFW uses all \gls{bfw} data train M (Fig.~\ref{fig:framework}). As mentioned, {we built \gls{bfw}} using data of VGG2, and there {is} no overlap {between} CASIA-Webface and LFW used to train the face encoder.

\subsection{De-bias experiment}
The percent error is a typical metric for \tipr{FR, as specialized figures (\eg plots and confusion matrices) are difficult for nontechnical audiences to interpret.} Specifically, global ratings (\eg average) are more practical to comprehend. {A prime example is to share the error rate per number {correctly predicted} (\eg falsely classify one in ten thousand).} For instance, claiming that a system predicts {an \gls{fp} in 1 of 10,000 predictions.} However, such an approximation can be hazardous, for it is inherently misleading. To {show} this, we ask the following questions. \emph{Does this hold for different demographics?} \emph{Does this rating depend on the faces - does it {carry} for all males and females?} {Setting} our system to {the} desired {\gls{fpr}} is fair regardless of population demographics (\ie subgroups).

The {questions above} were central to our previous work~\cite{robinson2020face}. We found the answer clear - {\emph{No, the reported \gls{fpr} is not true when analyzed per subgroup}.} {When} comparing the {\gls{fpr}} values (\ie the \emph{subgroup-specific} to the \emph{global}), the values drastically deviate from the \emph{global average} when the score threshold is fixed for all subgroups. Demographic-specific thresholds, meaning an assumption that demographic information is known prior to the problem, proved to mitigate the problem. However, prior knowledge of demographic, although plausible (\eg identifying a known subject on a \emph{blacklist}), {a strong assumption} limits the practical uses for which it could be deployed. To extend our prior work, we propose a de-biasing scheme to reduce the differences between the \emph{global} and \emph{subgroup-specific}. We set out to {claim} \emph{subgroup-specific} error rates to be fair across all involved demographics.

\glsunset{tpr} \glsunset{fpr}
\subsubsection{Metrics and settings}
{\gls{tpr}} and {\gls{fpr}} are used to examine the trade-off in confusion dependent on the choice of threshold discussed earlier. Specifically, we look at \emph{subgroup-specific} \gls{tpr} scores at the desired \gls{fpr}. We compute the following metric, the percent difference of the \emph{global} and \emph{subgroup-specific} {\gls{fpr}} values (\ie an average score is targeted) at a threshold \emph{l}. So we ask, ``\emph{How do the different subgroups compare to the average?}'' Specifically,

\begin{equation}
    \text{\% Error}(l) = (\frac{{\text{\gls{fpr}}(l)_{\text{subgroup}}} - {\text{\gls{fpr}}(l)_{\text{global}}}}{{\text{\gls{fpr}}(l)_{\text{global}}}})*100\%
\end{equation}

{The \emph{global results} are the results averaged across subgroups. {Then, the \emph{subgroup-specific results}, which differ meaningfully from the mean result (\ie the global results), are analyzed independently per subgroup.} Hence, there is a gap between \emph{global} and \emph{subgroup-specific}, which we show in Fig.~\ref{fig:page1-teaser-barplot} using the percent {error} (\ie $100\%*(\text{subgroup}-\text{global})/\text{global}$). Note that the percent {error} is negative when {global}$>${subgroup} (\ie \emph{subgroup-specific} are inferior).}

\subsubsection{Analysis}
The proposed balances the results while significantly boosting the {\gls{tpr}} at {\gls{fpr}.} The percent difference between {\emph{global} and \emph{subgroup-specific} \gls{fpr}} scores leads to {fairer} representation, especially at high FAR. Fig.~\ref{tab:ethnicy-far} shows the distribution of \gls{tpr} for the baseline (\emph{global}), proposed, and the optimal threshold (\emph{per subgroup}) at $\text{\gls{fpr}}=0.3$ (\ie the first column of the table above represented as a box-plot). Note that the standard deviation of \gls{tpr} using the baseline approach is {high, which we mitigate using the proposed scheme.} The proposed has thus boosted performance: {improved the rating and reduced the variances.} 

{We can interpret} Fig.~\ref{fig:page1-teaser-barplot} as a practical use case. A threshold is set to yield a specific \gls{fpr} (\ie how often an \gls{fp} is expected). The far-right (\ie 1e-4) claims 1 in 10,000 is incorrectly matched. Again, a verification system is set via a trade-off threshold (\ie $\theta$) that sets sensitivity: decreasing the score threshold {increases} the \gls{fpr} (Fig.~\ref{fig:metrics}). {Fig.~\ref{fig:page1-teaser-barplot} then compares the \emph{global} and \emph{subgroup-specific} on the subgroup level for a set of faces with equal representation for all subgroups.}
\begin{figure}[t!]
    \centering
    \begin{subfigure}[t]{1.25in}
        \includegraphics[width=\linewidth]{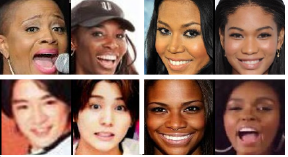}
        \caption{TN.}
        \label{fig:samples:a}
    \end{subfigure}
    \hspace{.1in}
    \begin{subfigure}[t]{1.25in}
        \includegraphics[width=\linewidth]{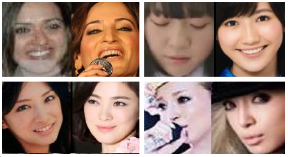}
        \caption{TP.}
         \label{fig:samples:b}
    \end{subfigure}
    \caption{{\textbf{Sample pairs.} Hard negatives (a) and positives (b) correctly matched with the proposed.}}
   \label{fig:hardpositives}
\end{figure}

The baseline shows a 200\% increase for WM at a \emph{global} \gls{fpr} of {1e-4, meaning} the \emph{global} expects an \gls{fp} per 10,000 \gls{tp}, while the \emph{subgroup-specific} performance doubles this (\ie all WM expect an \gls{fp} per 20,000 \gls{tp}).
	
{The} direction (\ie $\pm$) represents whether the difference is an improvement. A negative \%-difference {shows} a drop in performance compared to the \emph{global result}. For instance, AF with a -25\% difference for the baseline at an average \gls{fpr} of 1 in 10,000 implies that if the population of samples {comprises} only AF subjects, then the \gls{fpr} for the {chosen} $\theta$ for the claim of 1 in 10,000 would be 1 in 7,500. A consumer expecting a \gls{fpr} would only match this value when the sample population has the same distribution in samples per subgroup as the validation data for which $\theta$ was found.

{We can remove the percent difference} {via an} optimal threshold. {However, the assumption \tipr{is that} subgroup-specific thresholds can be determined from validation sets separated by subgroup. Also, the optimal solution assumes prior knowledge of the subgroup to which the sample of interest belongs {at test time.} Although the method was proof, \tipr{both assumptions} are impractical for most use cases. Hence, the proposed feature transformations reduce the percent differences from the original features (\ie the baseline). 

Fig.~\ref{fig:hardpositives} shows several hard positives {and negatives} incorrectly matched by the baseline but correctly identified by the proposed. These samples had scores closest to the global threshold (\ie score boundary). Notice the quality of at least one face per pair is low-resolution; extreme pose differences between the faces are also common. {The proposed scheme overcomes these challenges:} {mitigating bias boosts results,} and \tipr{several pairs change} from falsely being rejected to correctly being accepted.

\subsection{Privacy-preserving experiment} 
We aim to preserve identity information while de-biasing facial features, as {shown} in the prior experiment. We use a reverse gradient when training the subgroup branch to force the process to penalize the subgroup classifier when it is correct. Another benefit of the proposed de-biasing scheme is that it rids the facial {features} of demographic information, \tipr{which is useful for privacy and protection problems.} Ideally, face {features,} often the only representation of face information available at the system level, will not include attribute information like gender or ethnicity, as we prohibit the subgroup classifiers from learning. 

We train a {\gls{mlp}} to classify subgroups on top of the features to show how much subgroup information was removed. We can then measure the amount of information present in the face representation~\cite{acien2018measuring}.

The \gls{mlp} {comprises} three {fully connected ($\text{f}_\text{c}$)} layers {(\ie sizes 512, 512, and 256) and} the output {$\text{f}_\text{c}$ layer (\ie size 8, one per subgroup)} \tip{in Keras.} The first three layers were separated by ReLU activation and dropout~\cite{srivastava2014dropout} (\ie probability of 0.5), while only dropout (again, probability of 0.5) was {placed} \tipr{before} the output softmax layer. A categorical cross-entropy loss with Adam~\cite{kingma2014adam} set {with a  0.001 learning rate used to train.}

\begin{table}[t!]
\centering
    \caption{{\textbf{Subgroup classification results.} The baseline and proposed are on the left and right columns, respectively. Here, lower is better.}}
    \label{table:precrec}
  \resizebox{.85\linewidth}{!}{
 \begin{tabular}{rcccccccc}
     &\multicolumn{2}{c}{\cellcolor{blue!10}\textbf{Precision}}&& \multicolumn{2}{c}{\cellcolor{red!10}\textbf{Recall}}&& \multicolumn{2}{c}{\cellcolor{green!10}\textbf{F1}} \\
     \textbf{AF} &  0.962 &  0.734 &&  0.927 &  0.852&&0.943&0.788\tabularnewline
\textbf{AM} &  0.864 &  0.707 &&  0.974 &  0.730 &&0.915&	0.717\tabularnewline
\textbf{BF} &  0.940 &  0.655 &&  0.924 &  0.644 &&0.932&	0.647\tabularnewline
\textbf{BM} &  0.961 &  0.644 &&  0.962 &  0.668 &&0.961&	0.653\tabularnewline
\textbf{IF} &  0.961 &  0.641 & & 0.935 &  0.649 &&0.948&	0.644\tabularnewline
\textbf{IM} &  0.898 &  0.519 &&  0.902 &  0.589 &&0.898&	0.550\tabularnewline
\textbf{WF} &  0.934 &  0.554 &&  0.970 &  0.547 &&0.951&	0.549\tabularnewline
\textbf{WM} &  0.943 &  0.524 & & 0.848 &  0.317&&0.892&	0.392\\\midrule
\textbf{Avg.} &  0.933 &  0.622 &&  0.930 &  0.624 &&0.930&	0.617
\end{tabular}
}
\end{table}

\subsubsection{Metrics and settings}
We examine the accuracy of the subgroup classifiers via a confusion matrix. Specifically, we will look at how often each subgroup was predicted correctly and, when incorrect, the percentage it was mistaken for the others. The confusion was generated by averaging the five folds. Note the {top-performing} thresholds from the training folds on each test fold for the subgroup classifiers.

Also, we measure precision and recall. Precision is defined as $\text{P}(l)=\frac{\text{TP}}{\text{TP}+\text{FP}},$
which we average across subgroups $l\in L.$ The recall \tipr{(R)} is computed as
$\text{R}(l)=\frac{\text{TP}}{\text{TP}+\text{FN}}.$ This complements the confusion by allowing the specificity and sensitivity of the subgroups to be examined. There are inherent trade-offs between P and R. This motivates the $F_{1}$-score~\cite{jeni2013facing}, as the harmonic mean of P and R, $\text{F}_1 = 2*\frac{P*R}{P+R}.$

\subsubsection{Analysis}
{We showed the preservation} of identity knowledge (Fig.~\ref{tab:ethnicy-far}), and now we show the other benefits {of} privacy. The results {confirm} the {privacy-preserving} claim is accurate, leading to a 30\% drop in {predicting} gender and ethnicity from the {features} (Table~\ref{table:precrec}). 
Hence, {the} predictive power of all subgroups dropped significantly. {The} {decrease} in performance {suffices} to claim the predictions are now unreliable. Interestingly, it hindered the subgroups that the baseline {favored} the most from the de-bias scheme. {WM} and WF drop the most, while the AM and AF drop the least. {The} same trends in confusion propagate from the baseline to the proposed results (\eg WM mostly confuses {IM initially} and then again {with} the proposed). The same applies to cases of different sex.

\begin{figure}[!t]
\centering
\resizebox{\linewidth}{!}{
\newcommand\items{8}   
\arrayrulecolor{white} 
\noindent\begin{tabular}{c*{\items}{|E}|}
\multicolumn{1}{c}{} & 
\multicolumn{1}{c}{\textbf{AF}} & 
\multicolumn{1}{c}{\textbf{AM}} & 
\multicolumn{1}{c}{\textbf{BF}} &
\multicolumn{1}{c}{\textbf{BM}} & 
\multicolumn{1}{c}{\textbf{IF}} & 
\multicolumn{1}{c}{\textbf{IM}} &
\multicolumn{1}{c}{\textbf{WF}} & 
\multicolumn{1}{c}{\textbf{WM}} \\  \hhline{~*\items{|-}|} 
\textbf{AF} & 92.7 & 7.0  & 0.0  & 0.0  & 0.1  & 0.0  & 0.2  & 0.0  \\ \hhline{~*\items{|-}|}
\textbf{AM} & 1.6  & 97.4 & 0.1  & 0.1  & 0.0  & 0.2  & 0.0  & 0.7  \\ \hhline{~*\items{|-}|}
\textbf{BF} & 0.8  & 0.0  & 92.4 & 2.8  & 0.9  & 0.0  & 3.0  & 0.0  \\ \hhline{~*\items{|-}|}
\textbf{BM} & 0.0  & 0.0  & 2.0  & 96.2 & 0.0  & 1.6  & 0.0  & 0.2  \\ \hhline{~*\items{|-}|}
\textbf{IF} & 0.9  & 0.0  & 3.0  & 0.0  & 93.5 & 0.4  & 2.2  & 0.0  \\ \hhline{~*\items{|-}|}
\textbf{IM} & 0.0  & 3.6  & 0.0  & 0.8  & 2.0  & 90.2 & 0.0  & 3.3  \\ \hhline{~*\items{|-}|}
\textbf{WF} & 0.4  & 0.4  & 0.8  & 0.0  & 0.4  & 0.0  & 97.0 & 1.1  \\ \hhline{~*\items{|-}|}
\textbf{WM} & 0.0  & 4.6  & 0.0  & 0.1  & 0.3  & 8.6  & 1.6  & 84.8
\end{tabular}
\noindent\begin{tabular}{c*{\items}{|E}|}
\multicolumn{1}{c}{} & 
\multicolumn{1}{c}{\textbf{AF}} & 
\multicolumn{1}{c}{\textbf{AM}} & 
\multicolumn{1}{c}{\textbf{BF}} &
\multicolumn{1}{c}{\textbf{BM}} & 
\multicolumn{1}{c}{\textbf{IF}} & 
\multicolumn{1}{c}{\textbf{IM}} &
\multicolumn{1}{c}{\textbf{WF}} & 
\multicolumn{1}{c}{\textbf{WM}} \\  \hhline{~*\items{|-}|} 
& 85.2 & 8.8  & 1.1  & 0.3  & 1.7  & 0.6  & 1.5  & 1.0  \\ \hhline{~*\items{|-}|}
& 10.2 & 73.0 & 3.1  & 1.2  & 1.6  & 3.8  & 4.6  & 2.6  \\ \hhline{~*\items{|-}|}
& 3.5  & 3.8  & 64.4 & 11.4  & 5.4  & 3.2  & 6.5  & 1.8  \\ \hhline{~*\items{|-}|}
 & 1.5  & 2.1  & 13.3  & 66.8 & 4.5  & 5.7  & 4.0  & 2.1  \\ \hhline{~*\items{|-}|}
& 2.2  & 2.3  & 4.6  & 5.7  & 64.9 & 13.9  & 4.2  & 2.2  \\ \hhline{~*\items{|-}|}
& 1.7  & 3.7  & 1.4  & 5.4  & 9.9  & 58.9 & 8.4  & 10.6  \\ \hhline{~*\items{|-}|}
& 1.4  & 4.5  & 5.2  & 7.9  & 5.8  & 11.7  & 54.7 & 8.8  \\ \hhline{~*\items{|-}|}
& 10.4 & 5.4  & 6.1  & 6.0  & 7.9  & 17.1 & 15.4 & 31.7 
\end{tabular}
\arrayrulecolor{black} 
}
\caption{{\textbf{Subgroup confusion matrix.} Classification accuracy for the baseline (top) and proposed (bottom).}}
 \label{fig:confusion}
\end{figure}

Next, we examine the {confusion} for the different subgroups before and after de-biasing the face features (Fig.~\ref{fig:confusion}). As established, the baseline contains more subgroup knowledge, which a model can learn on top of. When trained and evaluated on \gls{bfw}, the baseline performs {best on F subgroups, which differs from the norm,} where M is {most} of the data. The WM is inferior in performance to all subgroups in either case.

\subsection{The privacy model}

To check the effectiveness of the proposed, we train M {on} the \gls{bfw} dataset and deploy {it} on the well-known LFW benchmark. We note that the training dataset we employ is significantly smaller than that used by \gls{soa} networks trained to achieve high performance on LFW using the MS1MV2 dataset, which contains 5.8 million images of 85,000 identities. {Even} though we initialize our network starting with features {learned} on MS1MV2, we train on a small dataset of 20,000 images of 800 subjects, two orders of magnitude smaller. The current \gls{soa} has 99.8\% verification accuracy. In comparison, the proposed scheme reaches its best score of 95.2\% after five epochs before dropping off and then leveling out around 81\% (Fig.~\ref{fig:lfwvalidation}). 
{The unbalanced data hinders the benefits of privacy and de-biasing} (\ie LFW {comprises} about 85\% WM). Furthermore, we optimized M by choosing the epoch with the best performance {before} the {drop-off.} Future steps could {improve} the proposed approach when transferring to unbalanced sets to detect the optimal settings.

\begin{figure}[t!]
    \centering
        \includegraphics[width=.6\linewidth]{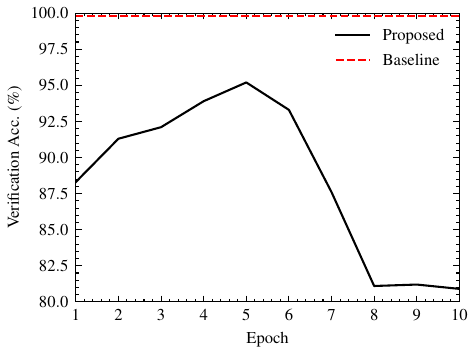}
    \caption{\textbf{Accuracy on LFW.} The proposed nears the {baseline performance} of 99.8\% with 95.2\% at Epoch 5, \tipr{preserving privacy with little} {accuracy diminished.}
    }
    \label{fig:lfwvalidation}
\end{figure}

\glsresetall	
\section{Conclusion}\label{sec:conclusions}
We show a bias for subgroups in \gls{fv} systems, where scores are converted to decisions via a predefined threshold. We previously introduced a \emph{subgroup-specific} threshold. We propose a novel approach: learn a lower-dimensional mapping that preserves identity and removes subgroup information, drawing inspiration from feature alignment. With the proposed method, the performance across subgroups balances and boosts accuracy. We reduce the difference between \emph{subgroup-specific} and \emph{global} performance across subgroups. Also, as knowledge of subgroups is removed from the features, privacy regarding demographics increases. 

The \gls{bfw} data and benchmarks address fairness in the data. Our feature encoder addresses privacy concerns by learning to map faces to a lower dimension that preserves identity and removes subgroup information. \gls{bfw} is at the forefront of ethical AI~\cite{nips2020ethics}.

The experimental settings and practices remain an open problem. For instance, gender labels are discrete values (\ie boolean): an approximation of sexuality best represented as real values~\cite{merler2019diversity}. Finer-grained or more specific subgroups could be another improvement (\eg Indians from North India versus South India, Black Africans versus African Americans, or distinguishing groups in Africa). We intend \gls{bfw} to serve as a benchmark for existing systems and a foundation for future researchers to extend.

{
\bibliographystyle{IEEEtran}
\bibliography{IEEEabrv,references}
}

\vspace{-10mm}

\begin{IEEEbiography}[{\includegraphics[width=1in,height=1.25in,clip,keepaspectratio]{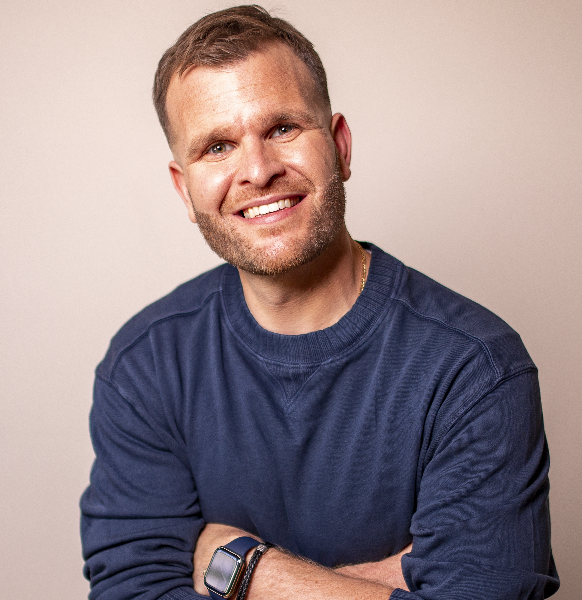}}]{Joseph P Robinson}
He earned a B.S. in ECE (2014) and a Ph.D. in CE (2020) from Northeastern University (NEU), where he also taught Data Analytics (2020\emph{Best Teacher}). Applied machine vision research emphasizes faces, deep learning, MM, and big data. Previously, led a team to TRECVid debut (MED, 3rd-place). Built many images and video datasets -- most notably FIW. Organized 2020 FG conference, various workshops, and challenges (\eg NECV, RFIW, AMFG, FacesMM), tutorials (MM, CVPR, FG), PC (\eg CVPR, FG, MIRP, MMEDIA, AAAI, ICCV), reviewer (\eg IEEE TBioCAS, TIP, TPAMI), and positions like President of IEEE@NEU (grad student adviser) and Relations Officer of IEEE SAC R1. Completed: NSF REUs (2010 \& 2011); interned at Analogic Corporation (2012) and BBN Tech. (2013), MIT Lincoln Labs (2014), System \& Tech. Research (2016 \& 2017), Snap Inc. (2018), and ISM (2019). Dr. Robinson worked at Vicarious Surgical as an AI Engineer, helping move public NYSE:RBOT (221-22); he taught ML at Tufts University (2023) while full-time at NEU. 
\end{IEEEbiography}

\vspace{40mm}
\begin{IEEEbiography}[{\includegraphics[width=1in,height=1.25in,clip,keepaspectratio]{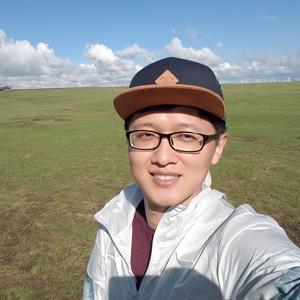}}]{Can Qin}
B.E. from the School of Microelectronics, Xidian University, China (2018), is pursuing a Ph.D. at the Department of Electrical and Computer Engineering, Northeastern University, under Dr. Yun Raymond Fu. His research focus spans transfer learning, semi-supervised learning, and deep learning in broad. He received awards for the \emph{Best Paper Award} of the Real-World Recognition from the \emph{Low-Quality Images and Videos} workshop at 2019 ICCV. Additionally, he has published at top-tier conferences (\ie NeurIPS, AAAI, ECCV). 

\end{IEEEbiography}

\begin{IEEEbiography}[{\includegraphics[width=1in,height=1.25in,clip,keepaspectratio]{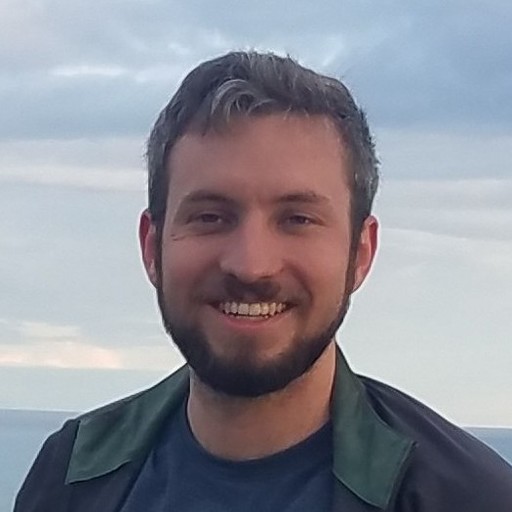}}]{Yann Henon} received a B.S. and an M.S from Monash University in Melbourne, Australia. His research has focused on medical imaging and has been published in top journals, including Developmental Cell, Journal of Applied Physiology and RSC Advances.
\end{IEEEbiography}

\begin{IEEEbiography}[{\includegraphics[width=1in,height=1.05in,clip,keepaspectratio]{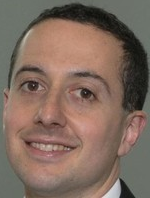}}]{Samson Timoner}
B.S. in Applied Physics from Caltech (1997); Ph.D. in Electrical Engineering and Computer Science from MIT (2003). He has organized numerous workshops, including CVPR View, CVPR ProCams, the New England CV Workshop, and the 4th RFIW. He has started two computer vision companies. He is an organizer of the computer vision community in Boston, bringing researchers and entrepreneurs together. He is currently leading the computer vision team at Wicket Software, focusing on facial detection and recognition algorithms.
\end{IEEEbiography}

\begin{IEEEbiography}[{\includegraphics[width=1in,height=1.25in,clip,keepaspectratio]{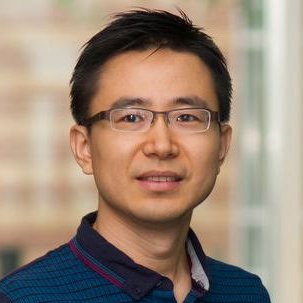}}]{Yun Fu}
(S’07-M’08-SM’11-F’19) received the B.Eng. degree in information engineering and the M.Eng. degree in pattern recognition and intelligence systems from Xi’an Jiaotong University, China, respectively, and the M.S. degree in statistics and the Ph.D. degree in electrical and computer engineering from the University of Illinois at Urbana-Champaign, respectively. He is an interdisciplinary faculty member affiliated with the College of Engineering and the College of Computer and Information Science at Northeastern University since 2012. His research interests are Machine Learning, Computational Intelligence, Big Data Mining, Computer Vision, Pattern Recognition, and Cyber-Physical Systems. He has extensive publications in leading journals, books/book chapters, and international conferences/workshops. He serves as associate editor, chairs, PC member and reviewer of many top journals and international conferences/workshops. He received seven Prestigious Young Investigator Awards from NAE, ONR, ARO, IEEE, INNS, UIUC, Grainger Foundation; eleven Best Paper Awards from IEEE, ACM, IAPR, SPIE, SIAM; many major Industrial Research Awards from Google, Samsung, Amazon, Konica Minolta, JP Morgan, Zebra, Adobe, and Mathworks, etc. He is currently an Associate Editor of the IEEE Transactions on Pattern Analysis and Machine Intelligence . He is a Fellow of AAAS, IEEE, IAPR, OSA and SPIE, a Lifetime Distinguished Member of ACM, Lifetime Member of AAAI, and Institute of Mathematical Statistics, member of Global Young Academy, INNS and Beckman Graduate Fellow during 2007-2008.
\end{IEEEbiography}

\end{document}